\documentclass[a4paper,fleqn]{cas-dc}

\usepackage[authoryear,longnamesfirst]{natbib}

\def\tsc#1{\csdef{#1}{\textsc{\lowercase{#1}}\xspace}}
\tsc{WGM}
\tsc{QE}
\tsc{EP}
\tsc{PMS}
\tsc{BEC}
\tsc{DE}

\begin{document}

\let\WriteBookmarks\relax
\def\floatpagepagefraction{1}
\def\textpagefraction{.001}

\shorttitle{MLN-net}

\shortauthors{Ke Wang et~al.}

\title [mode = title]{MLN-net: A multi-source medical image segmentation method for clustered microcalcifications using multiple layer normalization}                      

\tnotetext[1]{Corresponding authors.}

%

\author[1,2]{Ke Wang}[auid=000,bioid=1,
                    orcid=0000-0002-1926-4634]

\ead{wangke1992@zju.edu.cn}



\affiliation[1]{organization={College of Information Science and Technology},
    addressline={Zhejiang Shuren University}, 
    city={Hangzhou},
    country={China}}

\affiliation[2]{organization={State Key Laboratory of Industrial Control Technology},
    addressline={Zhejiang University}, 
    city={Hangzhou},
    country={China}}
\author[3]{Zanting Ye}[style=chinese]

\affiliation[3]{organization={School of Computer Science and Artificial Intelligence},
    addressline={Changzhou University}, 
    city={Changzhou},
    country={China}}

\author[4]{Xiang Xie}

\affiliation[4]{organization={School of Engineering},
    addressline={Newcastle university}, 
    city={Newcastle upon Tyne},
    country={UK}}

\author[5]{Haidong Cui}
\affiliation[5]{organization={Breast Surgery, First Affiliated Hospital},
    addressline={Zhejiang University}, 
    city={Hangzhou},
    country={China}}

\author[6]{Tao Chen}
\affiliation[6]{organization={The Second Affiliated Hospital of Zhejiang University School of Medicine},
    addressline={Zhejiang University}, 
    city={Hangzhou},
    country={China}}
    
\author[1]{Banteng Liu}[
   ]
\cormark[1]
\ead{3110102872@zju.edu.cn}







\begin{abstract}
Accurate segmentation of clustered microcalcifications in mammography is crucial for the diagnosis and treatment of breast cancer. Despite exhibiting expert-level accuracy, recent deep learning advancements in medical image segmentation provide insufficient contribution to practical applications, due to the domain shift resulting from differences in patient postures, individual gland density, and imaging modalities of mammography etc. In this paper, a novel framework named MLN-net, which can accurately segment multi-source images using only single source images, is proposed for clustered microcalcification segmentation. We first propose a source domain image augmentation method to generate multi-source images, leading to improved generalization. And a structure of multiple layer normalization (LN) layers is used to construct the segmentation network, which can be found efficient for clustered microcalcification segmentation in different domains. Additionally, a branch selection strategy is designed for measuring the similarity of the source domain data and the target domain data. To validate the proposed MLN-net, extensive analyses including ablation experiments are performed, comparison of 12 baseline methods. MLN-net enhances segmentation quality of full-field digital mammography (FFDM) and digital breast tomosynthe (DBT) images from the FFDM-DBT dataset, which achieves the average Dice similarity coefficient (DSC) of 86.52\% and the average Hausdorff distance (HD) of 20.49mm on the source domain DBT. And it outperforms the baseline models for the task in FFDM images from both the CBIS-DDSM and FFDM-DBT dataset, which achieves the average DSC of 50.78\% and the average HD of 35.12mm on the source domain CBIS-DDSM. Extensive experiments validate the effectiveness of MLN-net in segmenting clustered microcalcifications from different domains and the its segmentation accuracy surpasses state-of-the-art methods. Code will be available at https://github.com/yezanting/MLN-NET-VERSON1.
\end{abstract}





\maketitle

\section{Introduction}
\begin{figure}[!t]
\centerline{\includegraphics[width=1\linewidth]{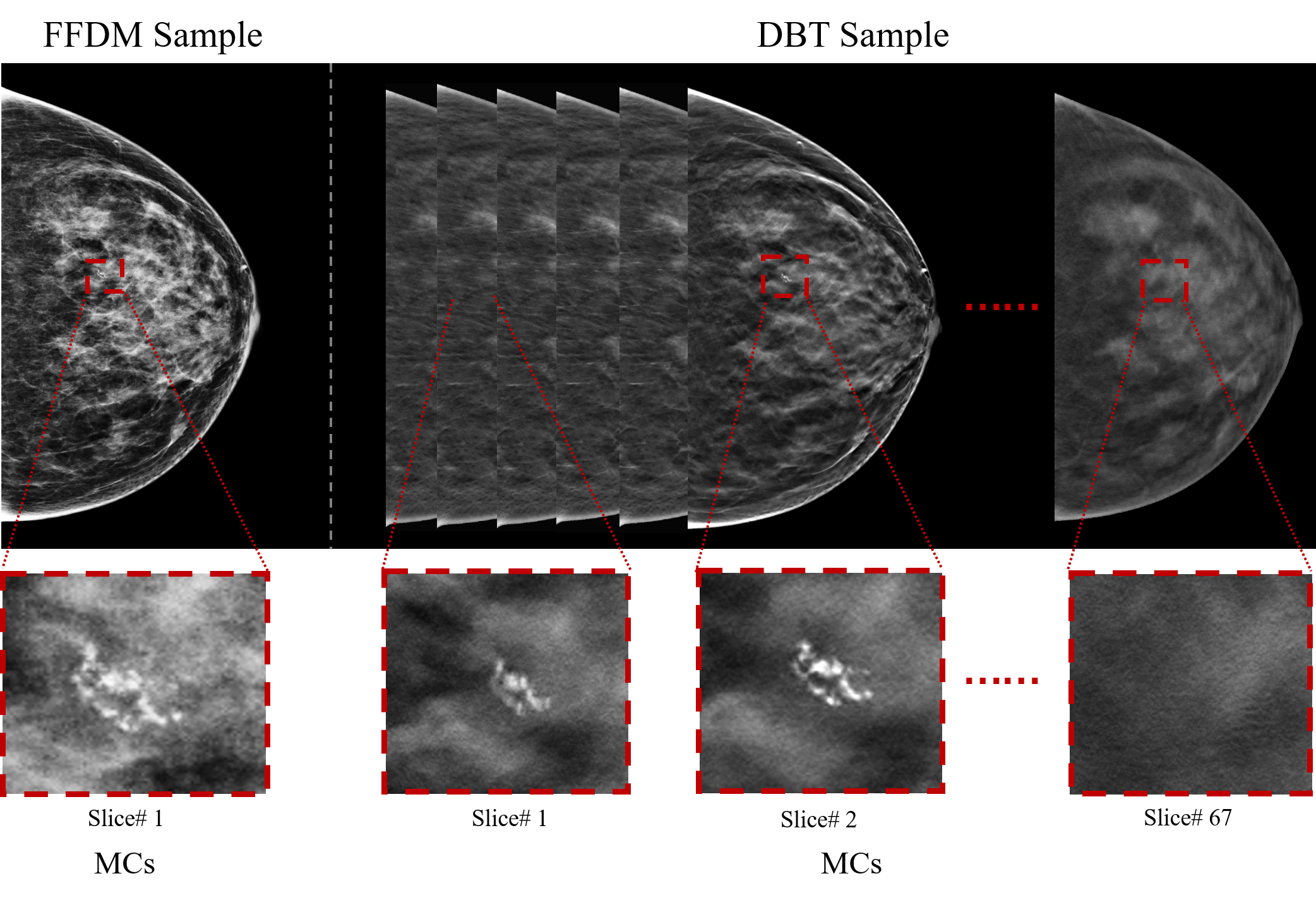}}
\caption{Examples of images from FFDM and DBT imaging modalities. The images are from the same patient and represent the same area of breast tissue. Clustered microcalcifications can be seen in both imaging modalities. Compared to FFDM, DBT performs slice photography of the lesion with a much narrower interval of 1mm, producing dozens to hundreds of images depending on individual differences.}
\label{fig1}
\end{figure}

Cancer is a significant public health concern worldwide. In 2020, there were an estimated 19.3 million new cancer cases, resulting in almost 10 million cancer-related deaths. Breast cancer is the most common form of cancer among women, accounting for 2.3 million cases and 11.7\% of all cancers worldwide \cite{sung2021global}. Available evidence proves that implementation of a screening program can decrease breast cancer mortality by up to 20\% \cite{mukama2020risk}. Therefore, improving the efficiency and accuracy of early breast cancer screening programs is essential \cite{youlden2012descriptive}. Mammography is widely used for screening breast cancer, with a detection rate ranging from 80\% to 90\%. And clustered microcalcifications, which appear in 30\% to 50\% of breast cancer patients, are among the primary pathological characteristics of breast cancer. The primary mammography modalities for screening clustered microcalcifications are FFDM and DBT \cite{tarver2012cancer,in2019facts}. Microcalcifications manifest as bright spots in both FFDM and DBT, as showns in Fig. \ref{fig1}. The shape, size, and distribution of clustered microcalcifications play a significant role in the diagnosis of benign and malignant breast cancer \cite{chong2019digital}. As such, improving the detection accuracy and efficiency of clustered microcalcifications in FFDM and DBT is an essential issue that requires.

Deep neural networks (DNNs) are widely recognized as the state-of-the-art technology for image analysis, providing superior performance in various tasks, including image classification, semantic segmentation, and object detection \cite{ibtehaz2020multiresunet,zhou2022generalizable,he2022evolutionary}. DNN-based computer-aided diagnosis (CAD) for clustered microcalcifications is rapidly and effectively being integrated into systems engineering. These methods use deep learning technology for automated screening of clustered microcalcifications in mammograms,
reducing the workload of medical professionals and enhancing the accuracy of diagnosis. For instance, Wang \emph{et al.} \cite{wang2018context} proposed a context-sensitive deep learning approach that utilized the local features of clustered microcalcifications as well as the surrounding tissue background to reduce false positives in the output.  Carneiro \emph{et al.} \cite{carneiro2017automated} proposed an automated methodology that combined information from cranio-caudal (CC) and medio-lateral oblique (MLO) views in FFDM and applied a deep learning model to extract the feature. The feature were then used to assess the risks of lesions. Zheng \emph{et al.} \cite{zheng20203d} proposed a novel 3D Context-Aware Convolutional Neural Network (CNN) for improving the accuracy of clustered microcalcifications detection in DBT. The method utilized a 2D CNN to extract the intra-slice features of DBT images, while a 3D CNN was employed to extract the inter-slice features. This method effectively employed multi-fault information in DBT to reduce false positives in the output. Samala \emph{et al.} \cite{samala2016deep} reconstructed the DBT volume using a multiscale bilateral filtering regularized simultaneous algebraic reconstruction technique, and subsequently used a CNN model to extract relevant features from the reconstructed volume. DNN-based methods assume that the source (training data) and target (test data) domains, as well as the the marginal distributions of the data, are supposed to be the same \cite{kowald2022transfer}, resulting in a close coupling of methods with data sources. While these DNN-based methods achieve high sensitivity in clustered microcalcifications detecting, they often face a reduction in performance when there is a substantial difference between the source and target domains. The phenomenon commonly referred to as the domain shift problem. Although a labelled dataset can help alleviate this problem, it can be costly for collecting such labeled medical datasets, and so these approaches are impractical.

Domain shifts frequently occur in clinical practice, presenting a significant obstacle to the effective use of DNNs. Factors such as the continuous advancement of image acquisition technology, variation in diagnostic procedures, diversity of scanners, and evolving imaging protocols may cause a decline in prediction accuracy on new data or render models obsolete due to these domain shifts \cite{perkonigg2021dynamic}. Our research focuses on identifying clustered microcalcifications in breast cancer, with a particular emphasis on countering the domain shift challenge in mammography. In mammography, gland density disparities among individuals, patient positioning, and exposure intensity variances on the same equipment contribute to intrinsic data variability. This issue becomes even more prominent when data is sourced from divergent imaging modalities (distinct imaging protocols), such as FFDM and DBT, representing extrinsic data variability.

Therefore, we propose MLN-net, which accepts single-domain data and can generalize to unseen domains. At its core, MLN-net seeks to tackle the domain shift issue, specifically when handling disparate clustered microcalcifications datasets. MLN-net has been meticulously designed to mitigate data variation impacts, enabling its application to new target datasets without necessitating retraining with labeled data. Furthermore, MLN-net is a medical image segmentation method, providing more detailed lesion information for the diagnosis and treatment of breast cancer.

Specifically, MLN-net reconstructs the original data exactly from the single-domain set to multi-domain under the Bézier curve and grayscale-inversion transformation, fostering greater data diversity and simulating potential shifts in the target domain. To perform multi-domain feature extraction, we propose a novel segmentation network with multiple LN layers. This network enhances domain information capture and reduces computation through the shared use of features among different inputs. Lastly, we propose a branch selection strategy, hinged on distance metric, for optimal LN layer selection within the segmentation network. During testing, this strategy gauges the distribution difference between the target and source domains by computing cosine similarity, thereby selecting the best segmentation results.

In summary, our contributions are summarized as follows:

\begin{itemize} \item We propose a novel framework for segmenting clustered microcalcificationsm, called MLN-net. MLN-net is capable of accurately segmenting multi-source images using only single-source images, thus effectively addressing the issue of domain shift between the source and target domains. \item We introduce a novel segmentation network with multiple LN layers to capture the feature of multi-source images. \item We develop a source domain data augmentation method based on Bézier curves and grayscale-inversion transformation to increase the diversity of the source domain data. Additionally, we develop a branch selection strategy to measure the similarities between the source domain data and the target domain data. \item We evaluate the proposed MLN-net on both the private FFDM-DBT dataset and the publicly available CBIS-DDSM dataset. MLN-net achieved superior performance over state-of-the-art methods. And in-depth analytical experiments demonstrate the efficacy of MLN-net.

\end{itemize}




\section{Related work}
\subsection{Domain shift}
Domain shifts arise from differences between the distribution of the training and the test data. In medical image analysis, these differences can result from different scanners, scanner generations, manufacturers, or imaging protocols during data acquisition \cite{perkonigg2021dynamic,he2021autoencoder}. For example, breast cancer diagnosis involves making use of medical image information obtained from both DBT and FFDM \cite{horvat2019calcifications,giess2017comparing}, corresponding to different imaging protocols. Even with the same imaging protocols, shown in Fig. \ref{fig7}, there can be significant discrepancies between images. These image disparities are a result of the rapid evolution of mammography technology. Therefore, to ensure successful deployment of deep learning-based clustered microcalcifications identification models in the changing environment, it is crucial to develop and advance methods that consider these domain shifts.

Recently, several methods have been proposed to address the issue of domain shift, encompassing semi-supervised transfer learning (STL) \cite{wang2021covid,abuduweili2021adaptive,zhang2021flexmatch,jakubovitz2019lautum,wei2019semi}, unsupervised domain adaptation (UDA) \cite{du2019ssf,dong2020cscl,doersch2015unsupervised}, and domain generalization (DG) \cite{fan2021adversarially,seo2020learning,zhang2020generalizing,zhou2020deep}. STL constructs effective self-supervised mechanisms for unlabeled data to migrates the rich knowledge of the source domain to the target domain. For instance, Abuduweili \emph{et al.} \cite{abuduweili2021adaptive} proposed an adaptive consistency regularization approach to leverage both pre-trained weights and unlabeled data. The adaptive consistency regularization approach consists of two complementary components: Adaptive Knowledge Consistency, which applies to the examples between the source and target models, and Adaptive Representation Consistency, which applies to the target model between labeled and unlabeled examples. Wang \emph{et al.} \cite{wang2021covid} utilized a pretrained model and a novel transfer feature learning model to extract image features, followed by proposing a deep fusion method and a selection method to combine the features from the pretrained and transfer feature learning model. This architecture enhanced the model's generalization capability. Unlike STL, UDA does not require a large amount of relevant data. UDA mitigates the distribution difference between the source and target domains to address the domain shift issue. For example, Du \emph{et al.} \cite{du2019ssf} and Dong \emph{et al.} \cite{dong2020cscl} applied deep adversarial networks to decrease the feature distribution gap among data from different domains so that the model can learn the same semantic features. Another noteworthy contribution includes \cite{he2021autoencoder}, which presented a system of three neural networks: the task model, the autoencoder model, and the adaptor model. The task model undertaken the image analysis task, while the autoencoder and adaptor models transfigured the features of the target domain to minimize domain shift. Other methods, such as those proposed by \cite{liu2020remove} and \cite{ma2019neural}, adopted style transfer methods to adapt the transformation of the target domain by analyzing the differences between the source and target domains.

STL and UDA face challenges when applied in the medical field due to privacy concerns that hinder the sharing of data between different hospitals and departments. These privacy concerns result in the target domain being unknown during the training process. Unlike STL and UDA, DG relies solely on the source domain data and builds models capable of directly generalizing to target domains, presenting a more pragmatic solution. Several DG methods have been proposed recently \cite{zhou2022generalizable,zhang2020generalizing,balaji2018metareg,dou2019domain,liu2020shape,fan2021adversarially,segu2023batch,seo2020learning,zhou2020deep}. Meta-learning, for instance, has been utilized to facilitate domain-invariant representation of multi-domain data \cite{balaji2018metareg,dou2019domain,liu2020shape}. There are also investigations that enhance the normalization process of DNNs by incorporating improved batch normalization (BN) and instance normalization (IN), thereby strengthening the networks' capacity to grasp domain information \cite{fan2021adversarially,segu2023batch,seo2020learning}. Additionally, strategies such as specific style transfer or data augmentation, which indirectly tackle potential changes in the target domain, have been successfully applied in medical image analysis to mitigate the issue of domain shift \cite{zhang2020generalizing,zhou2022generalizable}.

Clustered microcalcifications identification is a critical task in clinical practice \cite{pan2022molecular}. Existing DG methods are not directly applicable to the clustered microcalcifications segmentation task, mainly for two reasons. First, segmentation of the clustered microcalcifications involves distinction between the lesion's characteristics and the background, which presents a challenging classification problem. Second, most DG methods is relatively sensitive to small shifts of domain distribution and so is unable to provide a accurate segmentation to large domain shifts. To overcome these limitations, we propose a novel segmentation method for clustered microcalcifications that is the first segmentation method with domain generalization ability for clustered microcalcifications, filling a critical gap in the field.

\begin{figure*}
\centerline{\includegraphics[width=1\linewidth]{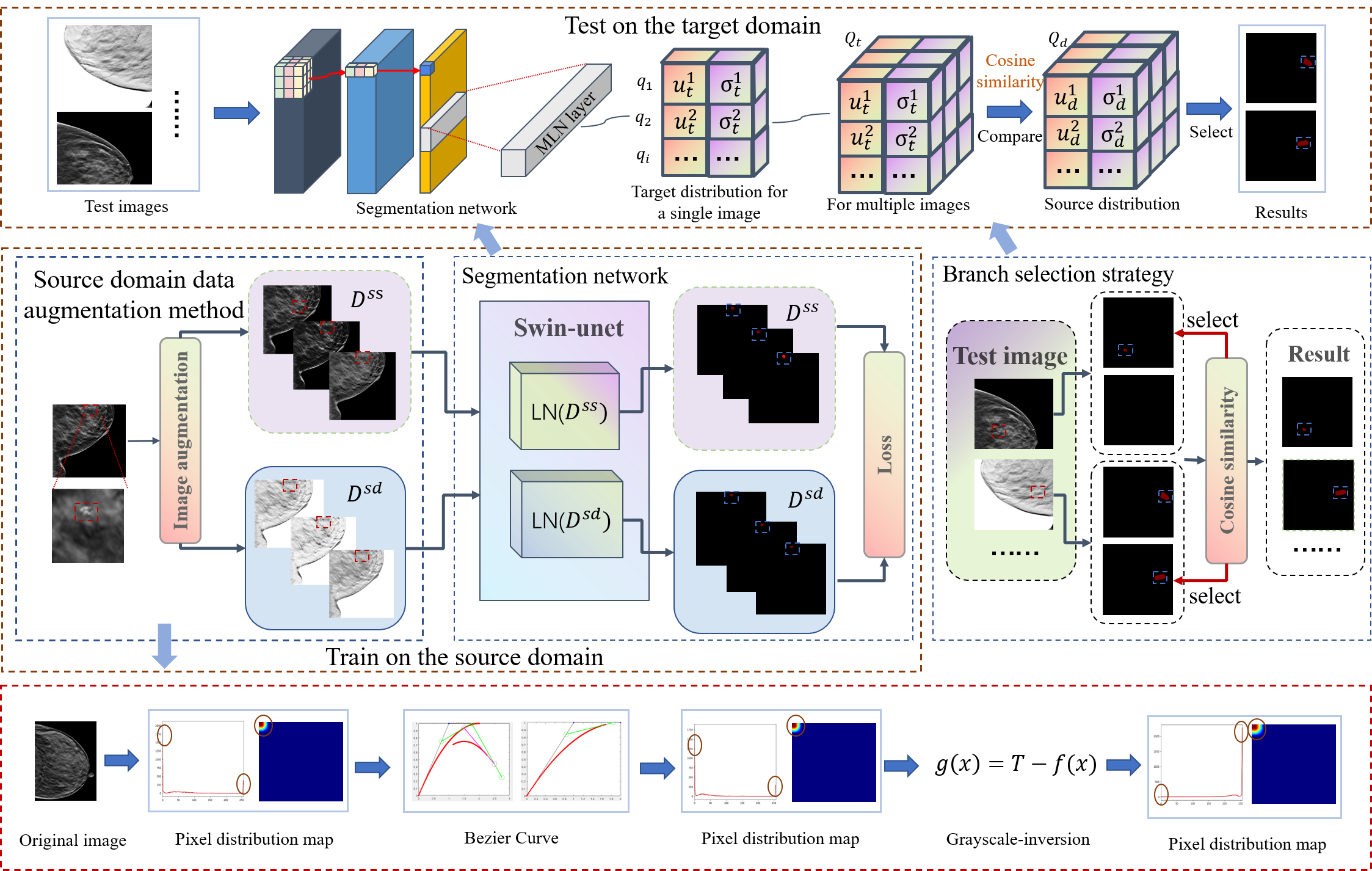}}
\caption{The schematic diagram of MLN-net framework. MLN-net utilizes source domain data for generalizable segmentation on unseen domain data. MLN-net is composed of a source domain data augmentation method, a segmentation network, and a branch selection strategy. During training stage, the source domain data augmentation method is utilized to augment the source domain data. The augmented data is then fed to the segmentation network with multiple LN layers, and the loss functions is used to optimize the segmentation results of different branches. During testing stage, the test data is fed into the trained segmentation network. And the branch selection strategy is adopted to choose the optimal segmentation results.}
\label{fig2}
\end{figure*}

\subsection{Medical image segmentation based on deep learning}

Medical image segmentation is a crucial component of automated medical image analysis, as it allows for the extraction of essential quantitative imaging markers, which in turn improves diagnosis, personalized treatment planning, and therapy monitoring \cite{sourati2019intelligent}. With the emergence of deep learning, segmentation approaches have evolved from traditional machine learning models to deep learning methods, yielding promising results in various segmentation tasks \cite{jiang2020deep}. CNNs \cite{he2016deep}, which utilize convolution and pooling operations to extract image features, has the widespread use in the medical imaging process. And there can be significant benefit in image classification and object detection of clustered microcalcifications \cite{bekker2015multi,wang2018context,zheng20203d,carneiro2017automated}. However, the image segmentation, as a pixel-level classification task, requires detailed image feature representation, which is often lost in image downsampling at pooling operations. Additionally, the deficiency of semantic information on shallow feature maps typically leads to poor segmentation performance for small targets with complex backgrounds.

Recently, vision transformer methods \cite{vaswani2017attention,cao2023swin,li2022can}, based on attention mechanism, have been introduced to exploit the correlations that exist in medical images between pixels. Compared with CNNs, these methods can preserve detailed features and show better feature representation capacity \cite{cao2023swin,chen2021transunet,bougourzi2023pdatt,gu2020net}. Representatively, Swinunet \cite{cao2023swin} and Transunet \cite{chen2021transunet} incorporated a self-attention mechanism into the basic Unet framework \cite{ronneberger2015u}, enhancing the global modeling ability and achieving excellent segmentation performance. PDAttunet \cite{bougourzi2023pdatt}, the latest work, which used the attention mechanism to segment the COVID-19 lesions. In our study, we employ Swinuet \cite{cao2023swin} as the backbone of segmentation network to extract image features. Diffident from the previous work \cite{cao2023swin}, we introduce a multiple LN layers structure into the network, thereby improving its capability to capture multi-domain information.

\section{Method}
Let ${{D}^{s}}=\sum\limits_{i}^{N}{\{x_{i}^{{}},y_{i}^{{}}\}} $ 
represents a source domain set, where \emph{S} denotes the domain label, $\emph{x}_i$ is the $\emph{i}_{th}$ image in the domain, $\emph{y}_i$ signifies the ground truth, and \emph{N} is the total number of samples within the domain. Our primary objective is to develop a multi-source image segmentation model ${{{M}_{\phi }}}$ that can achieve excellent generalization performance on source-similar domains ${{D}^{s\text{s}}}$ and the source-dissimilar domains ${{D}^{s\text{d}}}$. The segmentation model takes the form:

\begin{equation}
{{M}_{\phi }}(x)=y_{p}, (x) \in {{D}^{s}}^{s},{{D}^{s}}^{d},\ 
\label{eq1}
\end{equation}
\begin{equation}
\min (y_p-y_g)=0.\
\label{eq2}
\end{equation}
where $\phi $ represents the parameters of segmentation model, $\emph{x}$ denotes the input image, $\emph{y}_p$ signifies the mask of prediction, and $\emph{y}_g$ signifies the mask of ground truth.

The overall pipeline of MLN-net is illustrated in Fig. \ref{fig2}. MLN-net comprises three primary modules: a source domain data augmentation method, a segmentation network, and a branch selection strategy. This framework facilitates a generalizable segmentation approach that can extract features from source domain data and segment clustered microcalcifications on unseen domains. Subsequent sections will provide a detailed explanation of MLN-net's architecture.

\begin{figure}
\centerline{\includegraphics[width=1\linewidth]{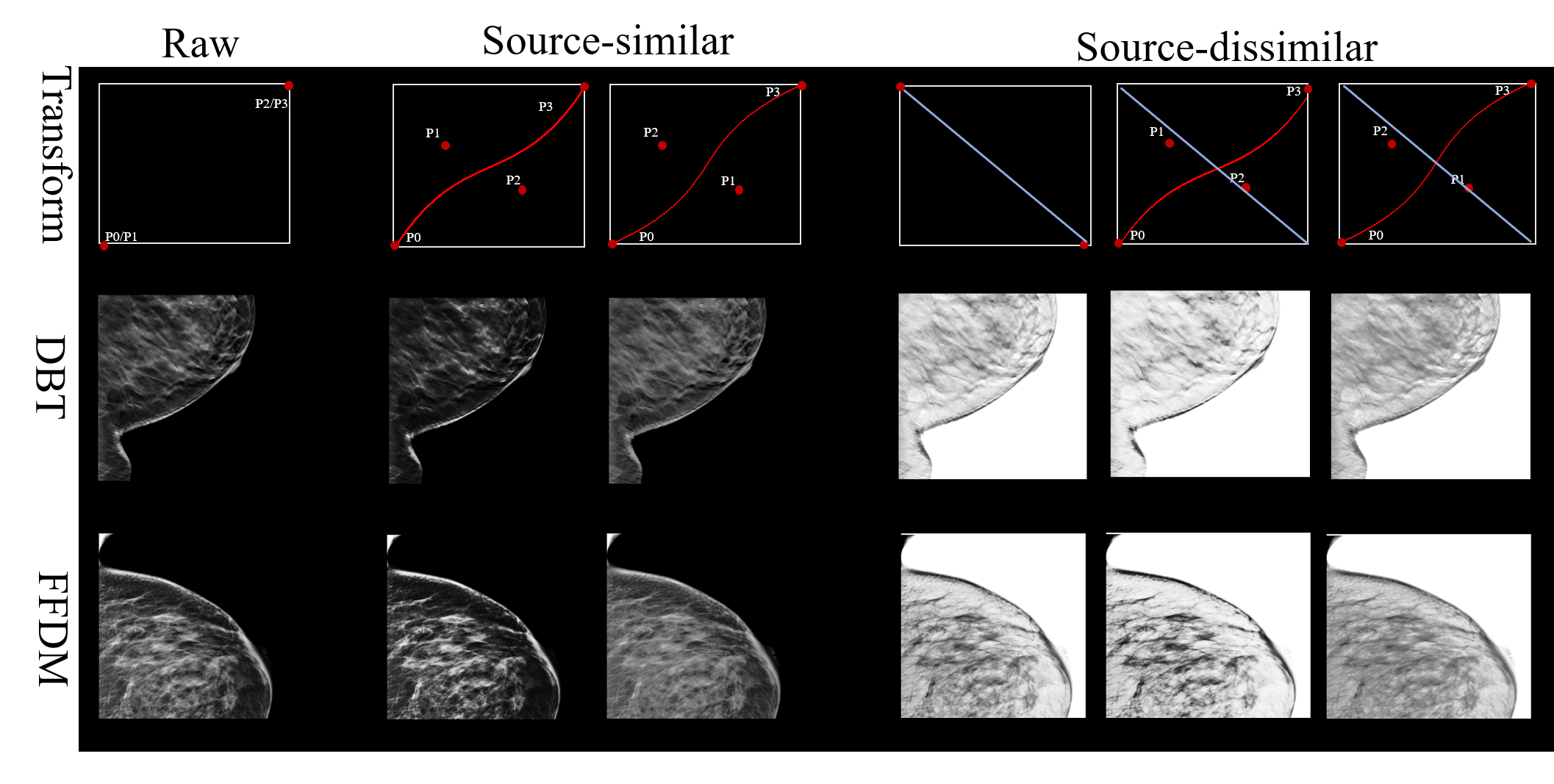}}
\caption{Examples of source-similar and source-dissimilar transformations results on FFDM-DBT dataset. In the first row, the red line signifies the transformation of source-similar based on the Bézier curve. This curve has four critical points; P1 denotes the starting point while P4 denotes the endpoint. P2 and P3 denote the control points that determine the curvature of the Bézier curve. In contrast, the blue line signifies transformation of source-dissimilar based on grayscale-inversion. And the second and the third row are the images after the transformations}
\label{fig3}
\end{figure}
    
\subsection{Source domain data augmentation}
Data augmentation method is frequently employed to improve accuracy in neural network recognition systems for medical data. And recent studies have suggested that data augmentation can alleviate the problem of domain shift \cite{zhang2020generalizing, zhou2022generalizable}. Motivated by these insights, we present, as depicted in Fig. \ref{fig3}, a monotonic non-linear transformation function that uses the Bézier curve to adjust pixel values, acquiring the source-similar data. Furthermore, we introduce a grayscale-inversion transformation method to acquire the source-dissimilar data.

\subsubsection{Source-similar data augmentation}
Mammography images, such as FFDM and DBT, are typically grayscale images. As depicted in Fig. \ref{fig1}, the distribution, boundary, and morphology of clustered microcalcifications in both types of images share similarities. The primary distinguishing factor is the pixel mapping strategy that defines light and dark features. Inspired by this observation, we propose a straightforward source-similar data augmentation method that utilizes the Bézier curve to transform the gray distribution of images. This method enhances the use of the source domain data and facilitates the generalization of the model to unseen target domains. The Bézier curve has a starting point, an end point, and two control points. Its mathematical expression can be represented as:
\begin{equation}
\ B(k)=\sum\nolimits_{i=0}^{N}{\left( \begin{matrix}
   N  \\
   i  \\
\end{matrix} \right){{P}_{i}}{{(1-k)}^{n-k}}{{k}^{i}}},N=3,k\in [0,1],\
\
\label{eq3}
\end{equation}

\begin{equation}
\ P=(a,b),a,b\in [0,1].\
\label{eq4}
\end{equation}
where $\emph{k}$ is a ratio of the length of the straight line, $\emph{P}$ represents the coordinate of the control points. The value of $\emph{P}$ is restricted to be between 0 and 1. In this study, we use two groups of curve control points, i.e., (0.30, 0.70) and (0.70, 0.30), as well as (0.70, 0.30) and (0.30, 0.70). And the starting point and end point of both groups are set to (0, 0), (1, 1). The setting of the above parameters will be discussed in detail in the ablation experiment.

\subsubsection{Source-dissimilar data augmentation}
In mammography images, lesions appear in the form of small white spots with black background, which results in decreased reliability in lesion identification. Besides, different imaging modalities apply varied pixel mapping approaches, further complicating the detection of pathological features. Standard DNN models that rely solely on superficial features may perform inadequately when presented with domain shifts. Because of the close similarity between the source-similar data, it is difficult to identify the corresponding internal feature mapping on domain shift tasks. Data augmentation method can therefore be used to give rise to variations in source-dissimilar data, in order to obtain a good generalization performance.

The grayscale-inversion transformation, as an image augmentation method, is commonly used in medical image processing to enhance the visibility of lesion areas. In our study, we introduce a novel use-case, employing the grayscale-inversion transformation, as a data augmentation method, to encourage an adaptive model. The training set is augmented using replicas of the training patterns, transformed according to the desired invariances. Specifically, anti-gray images, defined as source-dissimilar data, is introduced to allow a more complex target domain in the training dataset. This greatly reduces the dependence of the classification performance on notable features, which provides a framework for feature extraction with broad applicability. The mathematical expression of the grayscale-inversion transformation is presented as follows:
\begin{equation}
\ g(x,y)=T-f(x,y).\
\
\label{eq5}
\end{equation}
where $\emph{T}$ is the maximum pixel value of the image, $\emph{f}$ is the pixel value of the current position. $\emph{x}$, $\emph{y}$ are the coordinate positions, respectively.

\subsection{Segmentation network}
\subsubsection{Backbone of segmentation network}

\begin{figure}
\centerline{\includegraphics[width=0.92\linewidth]{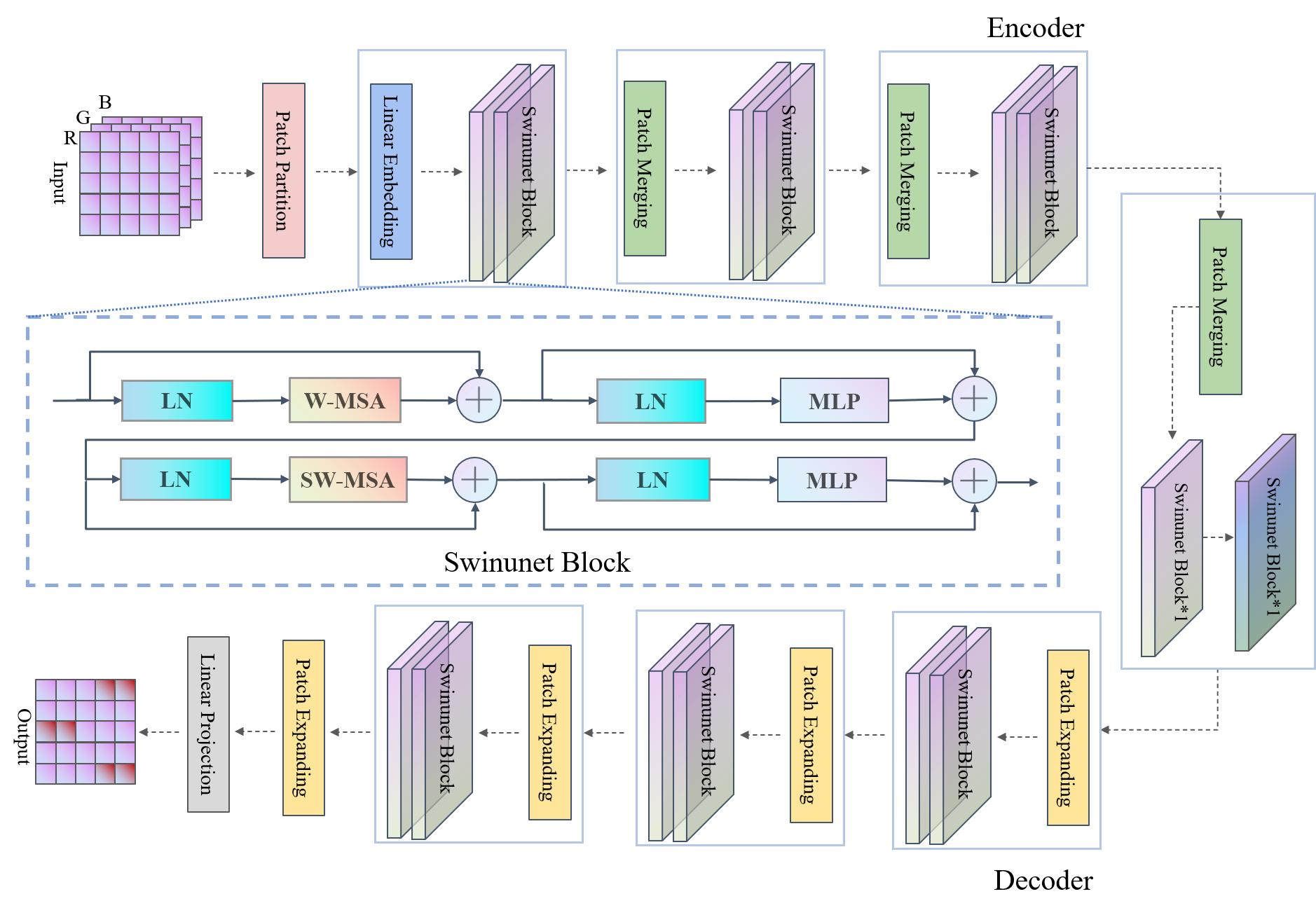}}
\caption{The composition structure of the backbone network and the Swinunet blocks in the Swinunet model.}
\label{fig4}
\end{figure}

In order to segment clustered microcalcifications across diverse domains, a novel segmentation network with multiple LN layers is introduced. This network utilizes Swin-Unet \cite{cao2023swin} as its backbone, with the self-attention mechanism for advanced feature representation learning. As depicted in Fig. \ref{fig4}, the Swin-Unet block is the heart of the Swin-Unet, primarily comprises of the window-based multi-head self-attention (W-MSA) and shifted window-based multi-head self-attention (SW-MSA) module. A window partitioning approach can therefore be proposed, which applies multi-head self-attention modules to two consecutive blocks. And successive Swin-Unet Transformer blocks based on this approach can be represented as follows:

\begin{equation}
\ {{\tilde{x}}^{t}}=W-MSA(LN({{x}^{t}}))+{{x}^{t-1}},\
\
\label{eq_expand1}
\end{equation}

\begin{equation}
\ {{x}^{t}}=MLP(LN({{\tilde{x}}^{t}}))+{{\tilde{x}}^{t}},\
\
\label{eq_expand2}
\end{equation}

\begin{equation}
\ {{\tilde{x}}^{t+1}}=SW-MSA(LN({{x}^{t}}))+{{x}^{t}},\
\
\label{eq_expand3}
\end{equation}

\begin{equation}
\ {{x}^{t+1}}=MLP(LN({{\tilde{x}}^{t+1}}))+{{\tilde{x}}^{t+1}}.\
\
\label{eq_expand4}
\end{equation}
where ${{x}^{t}}$ is the output of layer $t$. LN represents the LN layer. 

The self-attention mechanism of Swin-Unet can be viewed as performing a feature extraction. And information from such features can then be used to address long-term dependency issues prevalent in CNN-based methods \cite{zhang2022multi}. Unlike conventional self-attention based methods, Swin-Unet can capture the correlations between each region in the image and reduce computational overhead \cite{cao2023swin}.

\subsubsection{Multiple LN layers structure}
LN, a data normalization technique, can reduce the impact of internal covariance transformation and resolve the problem of vanishing and exploding gradients \cite{ba2016layer}. Unlike the commonly used BN method \cite{ioffe2015batch}, LN is insensitive to the input data batch size. This makes it an efficient normalization method in terms of self-attention based architectures. The mathematical expression of LN can be written as:

\begin{equation}
\ {{\text{u}}^{l}}=\frac{1}{n}\sum\limits_{i=1}^{n}{h_{i}^{l}},\
\
\label{eq6}
\end{equation}

\begin{equation}
\ {{\sigma }^{l}}=\sqrt{\frac{1}{n}\sum\limits_{i=1}^{n}{{{(h_{i}^{l}-{{\text{u}}^{l}})}^{2}}}},\
\
\label{eq7}
\end{equation}

\begin{equation}
\ {{L}^{t}}=\frac{\gamma }{{{\sigma }^{l}}}\odot ({{h}^{t}}-{{\text{u}}^{l}})+\omega. \
\
\label{eq8}
\end{equation}
where $h_{i}^{l}$ represents the $\emph{i}$ element of the $\emph{l}$ layer, \emph{n} represents the total number of elements at \emph{l} layer,  ${{\emph{u}}^{l}}$, ${{\sigma}^{l}}$ are mean and standard deviation; $\gamma $, $\omega $ are parameters for scaling and translation.

Data standardization is applied in LN to mitigate the variability in input data. However capturing domain distribution information from multiple domains poses a significant challenge. The deficiency of domain distribution information results in the overall loss incurred in decision making, which can lead to poor generalization and robustness. To resolve this issue, we introduce a segmentation network with multiple LN layers to extract the required distribution information of varying domains. The segmentation network utilizes the same backbone network parameters but different normalization strategies, which takes the form:
\begin{equation}
\ MLN(d)=\frac{{{\gamma }_{d}}}{\sqrt{\sigma _{d}^{2}+\Delta }}(x-{{\text{u}}_{d}})+{{\omega }_{d}}.\
\
\label{eq9}
\end{equation}
\\where \emph{d} is the label of domain, $\text{\emph{u}}$, $ \sigma $ are mean and standard deviation, $\Delta $ is a small constant value to avoid $\sqrt{\sigma _{d}^{2}+\Delta }=0 $, ${{\gamma }_{d}}$, ${{\omega }_{d}}$ are parameters for scaling and translation of \emph{d} domain.

During the training process, domain-specific information can be estimated by recording the parameters of the normalization layer for each domain at the final iteration. Evaluation of the best branch is straightforward as each normalization layer only involves the $\text{\emph{u}}$ and $\sigma$ values, which are recorded in training and test process. 

\subsubsection{Loss function}

MLN-net is modeled with multiple LN layers corresponding to a multi-branch structure. 
To evaluate the performance of all branches, we can formalize such issue through the introduction of a Dice loss, which is a overall measure of loss incurred in each branch. In this case, the overall loss can be written:

\begin{equation}
\ f_{dice}^{k}=-\frac{2}{N}\sum\limits_{n\in N}{\frac{\sum\nolimits_{i}{{{c}_{i,n}}{{z}_{i,n}}}}{\sum\nolimits_{i}{{{c}_{i,n}}+\sum\nolimits_{i}{{{z}_{i,n}}}}}},\
\
\label{eq_expand5}
\end{equation}

\begin{equation}
\ L=\sum\limits_{b=1}^{k}{f_{dice}^{b}}.\
\
\label{eq_expand6}
\end{equation}
where $k$ is the $k_{th}$ branch, $N$ is the total number of classification categories, ${c}_{i,n}$ is the softmax output that classifies the pixel $i$ into the class $n$ and ${z}_{i,n}$ is the ground truth of pixel $i$.

\subsection{Branch selection strategy}
The normalization process in segmentation network is expressed in terms of multiple LN layers, which leads to multiple outputs. And we can provide a much more general view of domain distribution information by giving a probabilistic interpretation to the multiple segmentation outputs \cite{segu2023batch}. A branch selection strategy, based on distance metrics, is developed, which calculate the distance between the mean and standard deviation of the target and source domain. Thus the domain distribution information is given by:
\begin{equation}
   \
   {{Q}_{d}}=[q_{d}^{1},q_{d}^{2},...,q_{d}^{l}] \
       =[(u_{d}^{1},\sigma _{d}^{1}),(u_{d}^{2},\sigma _{d}^{2}),...,(u_{d}^{l},\sigma _{d}^{l})], \
\
\label{eq10}
\end{equation}

\begin{equation}
   \
   {{Q}_{t}}=[q_{t}^{1},q_{t}^{2},...,q_{t}^{l}] \
       =[(u_{t}^{1},\sigma _{t}^{1}),(u_{t}^{2},\sigma _{t}^{2}),...,(u_{t}^{l},\sigma _{t}^{l})].\
\
\label{eq11}
\end{equation}
where \emph{d} represents the label of source domain, \emph{t} represents the label of target domain, and \emph{l} represents the $\emph{l}_{th}$ LN layer in MLN-net.


In practice, it is often feasible to use Euclidean distance metric for similarity evaluation. In branch selection strategy, we replace the traditional Euclidean distance by cosine distance to measure similarity between $Q_{d}$ and $Q_{t}$. Because cosine distance metric has superiority in quantifying similarity of high-dimensional space vectors \cite{strehl2000impact}. The corresponding strategy is given by:

\begin{equation}
   \
  C(Q_{d}^{i},Q_{t}^{i})=\frac{q_{d}^{i}\cdot q_{t}^{i}}{\left| q_{d}^{i} \right|\left| q_{t}^{i} \right|}, \
\
\label{eq12}
\end{equation}

\begin{equation}
   \
\text{Distance}({{Q}_{d}},{{Q}_{t}})=\sum\limits_{i=1}^{l}{C(}Q_{d}^{i},Q_{t}^{i}), \
\
\label{eq13}
\end{equation}

\begin{equation}
   \
S=\arg\min (\text{Distance}({{Q}_{d}},{{Q}_{t}})).\
\
\label{eq14}
\end{equation}
where \emph{d} represents the source domain, \emph{t} represents the target domain, \emph{i} represents the $\emph{i}_{th}$ LN layer of MLN-net. \emph{C} is cosine similarity.

\section{Experiments}
\subsection{Experiments setup}
\subsubsection{Data and pre-processing}
The private dataset FFDM-DBT and the public dataset CBIS-DDSM \cite{lee2017curated} serve to validate our methodology for segmenting clustered microcalcifications. The FFDM-DBT dataset consists of 80 patients’ DBT data, 420 patients’ FFDM data, each with corresponding pixel-level labels, sourced from the Breast Surgery department at the First Affiliated Hospital, Zhejiang University.
DBT data, each with a thickness of 1mm, incorporate double breast tomography images for every patient. The image count ranges from 70 to 200 per patient, maintaining a resolution of 1996×2457. Each patient's FFDM data includes four views: CC and MLO images of the left and right breasts, all at the same resolution of 1996×2457. Experienced radiologists manually annotated the labels for both FFDM and DBT data. The image preprocessing stage applies a 2×2 equal ratio segmentation on FFDM and DBT data to exclude the background blocks, and subsequently resamples the images to a 512×512 resolution, readying them for the source domain data augmentation method. For this study, we randomly selected 80\% of the data for the training set and reserved the remaining 20\% for the test set.

The CBIS-DDSM dataset includes 735 original breast images, each paired with corresponding pixel-level labels, procured from Massachusetts General Hospital, the University of South Florida, and Sandia National Laboratories. Similar to the FFDM-DBT dataset, data preprocessing procedures, such as image augmentation and resampling, are performed on the CBIS-DDSM dataset, thereby tailoring the data for our study.

\subsubsection{Evaluation metrics}
To evaluate the performance of MLN-net, we used five evaluation metrics: True Positive Rate (TPR), Precision (Pr), DSC, HD, and Average Surface Distance (ASD). TPR gauges the ability of MLN-net to detect lesions. Pr appraises its proficiency in identifying positive samples, whereas DSC assesses the extent of overlap between the predicted and actual annotations. The spatial overlap index for these three metrics (TPR, Pr, and DSC) ranges from 0 to 1, a higher value signifying superior segmentation performance. Conversely, HD and ASD evaluate the segmentation error, with lower values denoting superior segmentation performance. The metrics are defined as follows:

\begin{equation}
   \
\text{TPR}=\frac{\text{TP}}{\text{TP}+\text{FN}},\ 
\
\label{eq15}
\end{equation}

\begin{equation}
   \
\text{Pr}=\frac{\text{TP}}{\text{TP}+\text{FP}},\ 
\
\label{eq16}
\end{equation}

\begin{equation}
   \
\text{DSC}=\frac{\text{2TP}}{\text{2TP}+\text{FP}+\text{FN}},\ 
\
\label{eq17}
\end{equation}

\begin{equation}
   \
\text{ASD} = \frac{1}{n} \sum_{p\in P} \min_{g\in G} ||p-g||,\ 
\
\label{eq24}
\end{equation}

\begin{equation}
   \
\text{HD} = \max\left(\sup_{p\in P} \inf_{g\in G} ||p - g||, \sup_{g\in G} \inf_{p\in P} ||g - p||\right).\
\
\label{eq25}
\end{equation}
\\where TP is the number of true positives, TN is the number of true negative and FP is the number of false positives. \emph{P} and \emph{G} are the surface voxel set of predicted segmentation results and ground truth, respectively. \emph{s} and \emph{g} are an arbitrary voxel in \emph{P} and \emph{G}, and \emph{n} is the total number of \emph{S} and \emph{G} elements. $\Vert\cdot\Vert$ denotes the shortest Euclidean distance.

\subsubsection{Implementation details}
The implementation of the proposed MLN-net is based on Pytorch. The experimentation was conducted on a system operating with Windows 10 and equipped with an NVIDIA GeForce RTX 3080 graphics card, possessing 10GB memory. During training, the ADAM optimizer was employed with a batch size of 4, momentum set at 0.9, weight decay at 0.001, and a maximum of 100 epochs.

\subsection{Experiment \uppercase\expandafter{\romannumeral1}: Experimental results on FFDM-DBT dataset}
\begin{figure*}[!t]
\centerline{\includegraphics[width=1\linewidth]{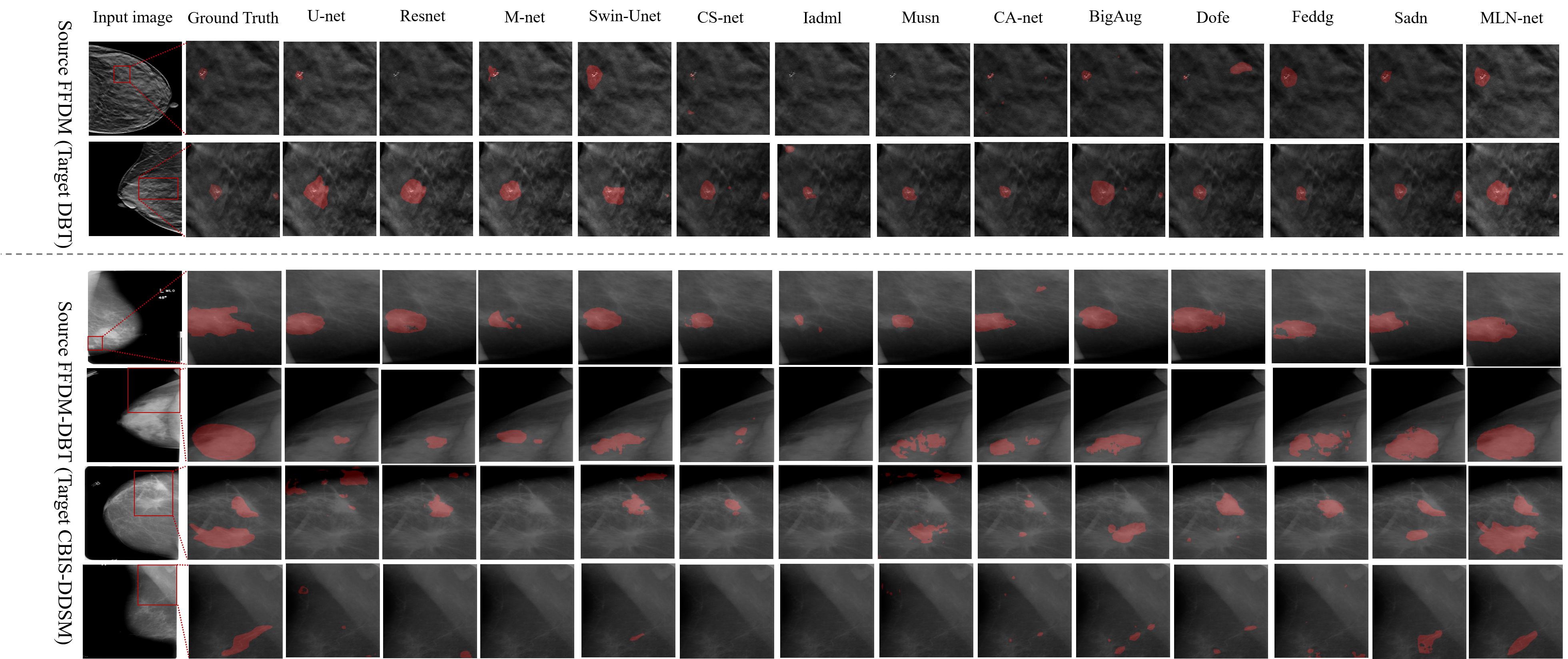}}
\caption{Clustered microcalcifications segmentation results of different methods. In the input and ground truth images, the red markings indicate the lesions annotated by doctors. The rest of the red markings represent the lesions identified by the different methods.}
\label{fig5}
\end{figure*}

\begin{figure*}[!t]
\centerline{\includegraphics[width=1\linewidth]{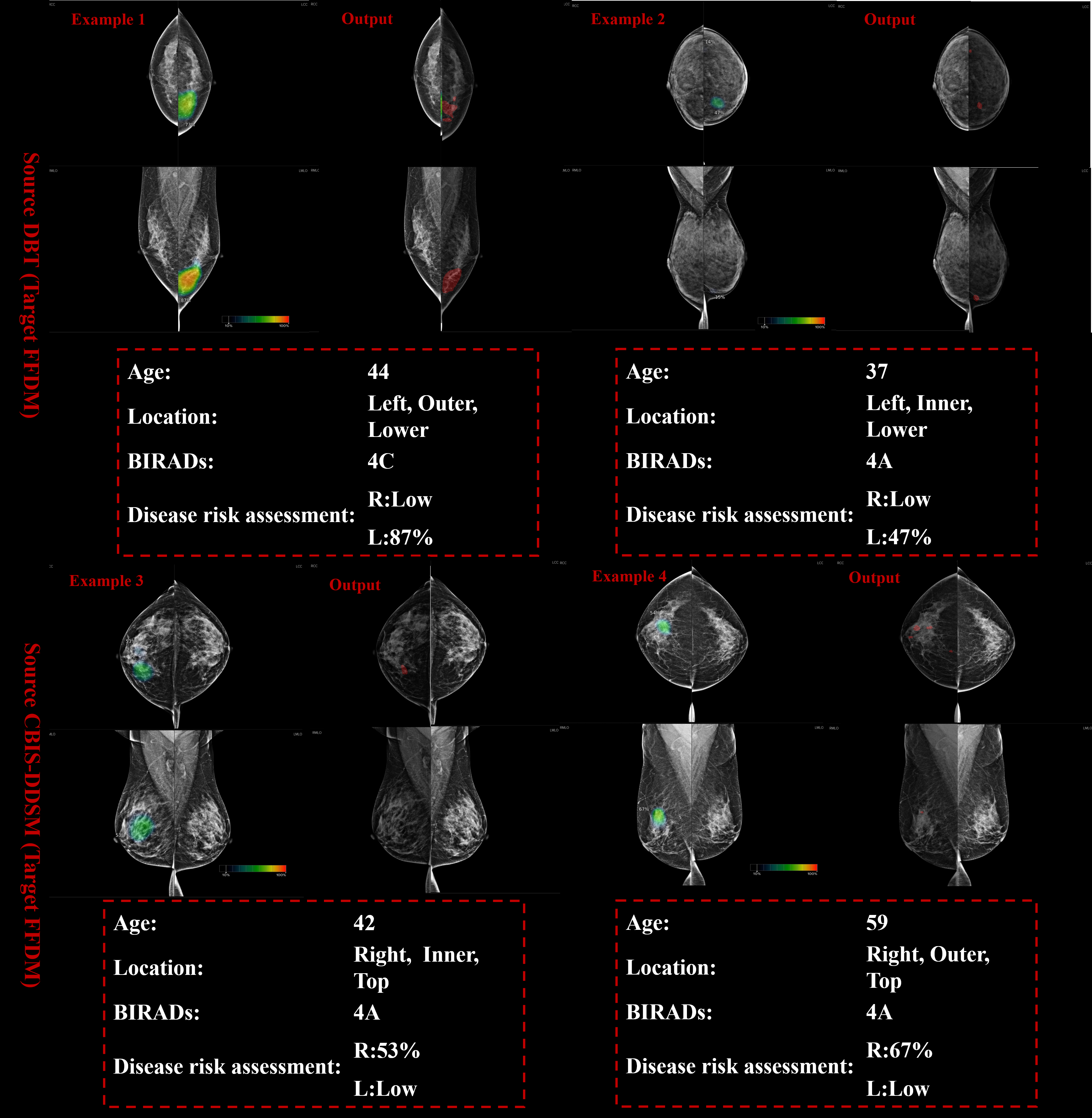}}
\caption{Example cases of MLN-net result on the target domain FFDM. This figure contains four cases, which trained on the different source domains. The left part of each case is the gold standard marked by doctors, and the right part is the output result of MLN-net. This figure shows the segmentation effectiveness of MLN-net on a complete case.}
\label{fig6}
\end{figure*}

In this section, the performance of the proposed MLN-net is assessed by taking experiments with the FFDM-DBT dataset, as shown in Table \ref{tab:1}. The results are segmented into three sections: P1 exhibits the performance of four basic segmentation networks; P2 exhibits the performance of four state-of-the-art methods for clustered microcalcifications recognition; P3 shows the performance of four state-of-the-art DG methods. Additionally, Fig. \ref{fig5} illustrates the segmentation results of MLN-net and other baseline methods on the target domain DBT. And Fig. \ref{fig6} visualizes the segmentation results of MLN-net alongside the physician-labeled gold standard for four comprehensive cases on the target domain FFDM, each case comprising four FFDM images from distinct imaging angles. A more detailed discussion of these results will follow.

\begin{table*}[cols=12,pos=h]
\caption{Segmentation performance comparison with twelve baseline methods on FFDM-DBT dataset. The baseline methods include P1: The basic segmentation methods, P2: The state-of-the-art methods for recognizing clustered microcalcifications and P3: The state-of-the-art DG methods. TPR, Pr, DSC, HD and ASD are used to evaluate the performance of these methods (best result is in bold for each column). Target DBT (Source FFDM): The models are trained on the domain FFDM and tested on the domain DBT, Target FFDM (Source DBT): The models are trained on the domain DBT and tested on the domain FFDM.}\label{tab:1}
\setlength{\tabcolsep}{1mm}{
 \begin{tabular}{cp{4.6em}cccccccccccc}
 \toprule
 \multicolumn{2}{c}{} &       & \multicolumn{5}{c}{Target DBT (Source FFDM)} &       & \multicolumn{5}{c}{Target FFDM (Source DBT)} \\
\cmidrule{1-2}\cmidrule{4-8}\cmidrule{10-14}          & Method        &       &TPR (\%) & Pr (\%) & DSC (\%) & HD (mm) & ASD (mm)&       & TPR (\%) & Pr (\%) & DSC (\%) & HD (mm) & ASD (mm) \\
\cmidrule{1-2}\cmidrule{4-8}\cmidrule{10-14}    \multicolumn{1}{c}{\multirow{4}[2]{*}{P1}} & Unet   &       & 48.52 & 35.77 & 37.32 & 51.97 & 20.65 &       & 52.32 & 36.74 & 43.63 & 44.64 & 21.91 \\
          & Resnet   &       & 46.45 & 39.16 & 37.25 & 42.05 & 16.76 &       & 46.07 & 47.46 & 46.98 & 35.32 & 17.06 \\
          & M-net  &       & 57.13 & 60.22 & 55.56 & 30.47 & 13.10  &       & 52.71 & 65.97 & 59.41 & 25.98 & 13.76 \\
          & Swin-Unet &       & 50.09 & 46.41 & 46.97 & 29.17 & 13.78 &       & 53.65 & 51.80 & 57.07 & 26.21 & 13.74 \\
\cmidrule{1-2}\cmidrule{4-8}\cmidrule{10-14}    \multicolumn{1}{c}{\multirow{4}[2]{*}{P2}} & CS-net &       & 65.42 & 68.81 & 66.28 & 28.86 & 10.99 &       & 67.21 & 69.65 & 67.72 & 28.12 & 11.08 \\
          & Iadml &       & 52.24 & 55.82 & 47.67 & 67.91 & 29.64 &       & 48.07 & 50.32 & 47.64 & 56.42 & 27.03 \\
          & Musn  &       & 57.19  & 55.62 & 55.32 & 45.18 & 22.63 &       & 61.27 & 59.74 & 58.49 & 39.98 & 19.12 \\
          & CA-net &       & 71.15 & 72.40  & 77.92 & 26.50  & 8.71  &       & 73.06 & 77.59 & 78.06 & 22.03 & 9.59 \\
\cmidrule{1-2}\cmidrule{4-8}\cmidrule{10-14}    \multicolumn{1}{c}{\multirow{4}[2]{*}{P3}} & BigAug &       & 70.91 & 79.66 & 76.96 & 29.15 & 12.01 &       & 71.70  & 76.43 & 74.26 & 26.74 & 11.85 \\
          & Dofe  &       & 65.58  & 59.96 & 59.77 & 37.47 & 18.14 &       & 67.57 & 66.49 & 66.58 & 34.40  & 16.01 \\
          & Feddg &       & 73.59 & \textbf{82.74} & 75.33 & 31.48 & 12.97 &       & 77.11  & 79.70  & 77.79 & 26.24 & 9.94 \\
          & Sadn &       & \textbf{79.36} & 81.97 & 74.95 & 27.78 & 9.20   &       & 83.06 & 78.84 & \textbf{87.25} & 23.80  & 7.67 \\
\cmidrule{1-2}\cmidrule{4-8}\cmidrule{10-14}          & MLN-net &       & 78.43 & 81.02 & \textbf{78.91} & \textbf{23.32} & \textbf{6.72} &       & \textbf{85.49} & \textbf{88.75} & 86.52 & \textbf{20.49} & \textbf{5.96} \\
 \toprule
\end{tabular}}
\end{table*}

\subsubsection{Comparison with the basic segmentation methods}

The Table \ref{tab:1} (P1) shows the results of four basic medical image segmentation methods. Among these methods, both Unet \cite{ronneberger2015u} and Resnet \cite{he2016deep} are fully convolutional neural network (FCNN) based methods. On the source domain FFDM, Unet and Resnet achieved TPR of 48.52\% and 46.45\%, as well as DSC of 37.32\% and 37.25\%, respectively. Resnet exhibites a 3.39\% improvement in Pr over Unet.  Since both Unet and Resnet have similar feature extraction processes, they yield comparable results. M-net \cite{fu2018joint} and Unet employ a U-shaped network, but M-net introduces a novel segmentation mode with multi-scale input and side output. M-net constructs multi-level receptive fields and develops multi-distance dependence relationships to enhance predictive performance. The use of diverse scale receptive fields is beneficial in addressing the issue of image feature extraction. Furthermore, M-net utilizes side-output layers and multi-label loss functions to capture both global and local characteristics, respectively. These proposed loss functions establish an information correlation between local lesions and global images, enhancing the segmentation power of unseen and variable target images. And on the source domain FFDM, M-net significantly enhances TPR, Pr and DSC by 8.61\%, 24.45\% and 18.24\%, respectively, compared to Unet, confirming their beneficial effect on domain shift problem. However, Unet, Resnet, and M-net exhibit a low overall accuracy due to the FCNN's limited ability to learn explicit global and long-term semantic information interactions. This constraint directly affects their cross-center generalization ability. Moreover, FCNN's pooling operation can cause the loss of detailed features, leading to low recognition rates of small and discrete clustered microcalcifications. Swin-Unet \cite{cao2023swin} leverages a U-like segmentation network that employs local window self-attention instead of convolution and pooling operations. And it employs fixed window self-attention and window self-attention sliding strategy, which results in deep feature learning while reducing computational cost. The effectiveness of Swin-Unet has proven in clustered microcalcifications segmentation tasks, achieving TPR of 53.65\% , Pr of 51.80\% , and DSC of 57.07\% on the source domain DBT. 

Swin-Unet is specifically designed for solving lesion segmentation within a single domain. Comparison to CNNs, Swin-Unet demonstrates potential in addressing domain shift issues, which can be attributed to its ability to incorporate long-term semantic information interactions \cite{kirillov2023segment,butoi2023universeg}. In addition, every approach demonstrates an appreciable performance enhancement when utilizing DBT as the source domain, in comparison to FFDM. This enhancement can likely be attributed to the wide array of tomographic images encompassed in the DBT data. These data are notably more comprehensive and encompass a broader diversity of lesion shapes and contours.

\subsubsection{Comparison with the state-of-the-art methods for recognizing Clustered microcalcifications}
In this section, MLN-net is contrasted with the state-of-the-art methods for recognizing clustered microcalcifications \cite{hossain2022microc,zheng20203d,wang2018context,wichakam2018deep}. And the quantitative results are presented in Table \ref{tab:1} (P2). CS-net \cite{wang2018context} and Musn  \cite{hossain2022microc} are developed on FFDM data. CS-net employs two parallel subnetworks to detect lesions in images of varying resolutions, effectively preserving intricate details. In our experiment, we selected images with 512x512 and 1024x1024 resolutions as CS-net's input. On the source domain FFDM, CS-net achieves TPR of 65.42\% and DSC of 66.28\%. These results underscore the potential advantages of multi-scale inputs in capturing domain-specific information, a finding consistent with that of M-net. Musn augments FFDM data via a local Laplacian filter and C-means clustering algorithm to select the region of interest, followed by a modified Unet for segmentation. However, Musn's excessive emphasis on local image information limits its ability to process global data, which produces a significant reduction in generalization, resulting in the subpar DSC of 45.18\% on the FFDM source domain. Iadml \cite{wichakam2018deep} and CA-net \cite{zheng20203d}, developed on DBT data, use CNN architectures to detecte clustered microcalcifications. Iadml merely uses four convolutional layers for feature extraction, which consequently leads to a potential loss of deep semantic features. In comparison, CA-net constructs a step-wise screening detection model, drawing upon a deeper CNN architecture and a more intricate detection process. It achieves DSC of 77.92\% on the FFDM source domain even domain shift occur. In our implementation of CA-net, we replicated 2D FFDM and DBT images to bestow a third dimension, as CA-net applies 3D convolution for image feature extraction. This transformation, which can be perceived as a form of data augmentation, could potentially improve CA-Net's performance, leading to an overestimation of its capability.

Current deep learning methods applied in clustered microcalcifications detection are originally designed for data on the same domain, thereby overlooking the challenge of domain shift. The swift advancement of medical imaging technology, which leads to data transformation, has profoundly curtailed the applicability of these techniques. To remedy this shortcoming, we propose MLN-net: A recognition method for clustered microcalcifications boasting cross-domain generalization capabilities. This method alleviates the intrinsic dependence of deep learning on substantial data, and holds the potential to surmount the data barriers spanning diverse imaging devices.

\subsubsection{Comparison with the state-of-the-art DG methods}
In this section, four state-of-the-art DG methods are chosen for comparsion, including BigAug \cite{zhang2020generalizing}, Dofe \cite{wang2020dofe}, Feddg \cite{liu2021feddg} and Sadn \cite{zhou2022generalizable}. BigAug, using a deep stacked transformation approach, emulates domain shifts for specific medical imaging modalities by augmenting data on a solitary source domain. Notably, on the source domain FFDM, BigAug achieves DSC, HD, and ASD of 74.26\%, 26.74mm, and 11.85mm, respectively, underscoring the potency of simple data augmentation in addressing domain shift issues. Dofe, a domain-invariant feature learning method, leverages multi-source domain knowledge via the proposed domain knowledge pool to improve the generalization ability. To ensure a fair comparison between MLN-net and Dofe, we used both source-similar and source-dissimilar images as inputs to Dofe. However, Dofe's segmentation performance proves subpar, achieving DSC of 59.77\% and HD of 18.14mm on the source domain FFDM. This diminished performance may result from Dofe's design specificity for Fundus image segmentation tasks, where its extensive prior knowledge may obstruct generalization to different segmentation tasks. Feddg, a mata-learning-based approach, demonstrates performance comparable to BigAug. Unlike BigAug, Dofe, and Feddg, which focus on learning or keeping domain-invariant information, Sadn is combined with data augmentation methods and make use of information about similar domain to improve generalization. This results in an improvement in the DSC by approximately 10\% compared to Feddg on the source domain DBT. The success of Sadn may arises from recognizing similarities of various medical imaging technologies, where the variance is considerably less than that between different imaging technologies used for natural images.

Inspired by Sadn, MLN-net combines multiple LN layers structure with branch selection strategy and thereby selects the optimal network branch. And by the introduction of self-attention mechanism, we can obtain a significant further improvement in feature extraction. Experinment results shows that MLN-net gives better performance on evaluation metrics and outperforms other approaches in HD and ASD by a significant margin.

\begin{table*}[cols=12,pos=h]
\caption{Segmentation performance comparison with twelve baseline methods on FFDM-DBT and CBIS-DDSM dataset. The baseline methods include P1: The basic segmentation methods, P2: The state-of-the-art methods for recognizing clustered microcalcifications and P3: The state-of-the-art DG methods. TPR, Pr, DSC, HD and ASD are used to evaluate the performance of these methods (best result is in bold for each column). Target FFDM-DBT (Source CBIS-DDSM): The models are trained on the domain CBIS-DDSM and tested on the domain FFDM-DBT, Target CBIS-DDSM (Source FFDM-DBT): The models are trained on the domain FFDM-DBT and tested on the domain CBIS-DDSM.}\label{tab:2}
\setlength{\tabcolsep}{1mm}{
 \begin{tabular}{cp{4.6em}cccccccccccc}
 \toprule
    \multicolumn{2}{c}{} &       & \multicolumn{5}{c}{Target FFDM-DBT (Source CBIS-DDSM)} &       & \multicolumn{5}{c}{Target CBIS-DDSM (Source FFDM-DBT)} \\
\cmidrule{1-2}\cmidrule{4-8}\cmidrule{10-14}          & Method &       & TPR (\%) & Pr (\%) & DSC (\%) & HD (mm) & ASD (mm) &       & TPR (\%) &Pr (\%) & DSC (\%)& HD (mm) & ASD (mm) \\
\cmidrule{1-2}\cmidrule{4-8}\cmidrule{10-14}    \multicolumn{1}{c}{\multirow{4}[2]{*}{P1}} & Unet   &       & 27.62  & 29.39  & 26.97  & 50.92  & 36.69  &       & 22.14  & 24.08  & 19.92  & 52.51  & 38.16  \\
          & Resnet &       & 24.76  & 27.15  & 27.36  & 58.36  & 47.05  &       & 19.54  & 22.28  & 21.87  & 50.33  & 38.80  \\
          & M-net &       & 9.87  & 13.19  & 12.32  & 69.90  & 44.18  &       & 7.67  & 8.91  & 6.72  & 71.77  & 47.05  \\
          & Swin-Unet &       & 33.39  & 34.29  & 35.10  & 31.25  & 34.75  &       & 37.81  & 30.15  & 29.97  & 46.04  & 30.58  \\
\cmidrule{1-2}\cmidrule{4-8}\cmidrule{10-14}    \multicolumn{1}{c}{\multirow{4}[2]{*}{P2}} & CS-net &       & 28.71  & 26.94  & 23.19  & 45.16  & 32.63  &       & 24.45  & 19.80  & 20.07  & 51.37  & 35.87  \\
          & Iadml&       & 7.61  & 10.07  & 8.10  & 77.48  & 52.66  &       & 8.23  & 8.68  & 6.29  & 81.78  & 59.15  \\
          & Musn &   & 16.48  & 20.73  & 21.01  & 55.50  & 41.36  &       & 20.12  & 22.77  & 18.56  & 55.14  & 40.70  \\
          & CA-net &       & 35.13  & 27.93  & 33.67  & 39.97  & 32.10  &       & 30.16  & 26.76  & 25.82  & 56.18  & 35.75  \\
\cmidrule{1-2}\cmidrule{4-8}\cmidrule{10-14}    \multicolumn{1}{c}{\multirow{4}[2]{*}{P3}} & BigAug  &  & 55.14  & \textbf{56.39} & 44.51  & \textbf{33.64} & 22.58  &  & 41.93  & 43.17  & 40.48  & 39.99  & 25.03  \\
          & Dofe  &  & 32.76  & 34.83  & 29.48  & 55.79  & 31.75  &   & 37.16  & 33.87  & 31.05  & 46.74  & 28.10  \\
          & Feddg &  & 43.08  & 41.32  & 37.97  & 48.95  & 29.84  &   & 37.15  & 37.32  & 38.82  & 46.04  & 26.99  \\
          & Sadn   &   & 51.75  & 49.92  & 45.16  & 41.04  & 26.11  &   & 39.49  & 41.60  & 42.35  & 48.00  & 26.72  \\
\cmidrule{1-8}\cmidrule{10-14}          & MLN-net &       & \textbf{56.85} & 54.57  & \textbf{50.78} & 35.12  & \textbf{20.33} &       & \textbf{48.35} & \textbf{49.41} & \textbf{45.76} & \textbf{36.74} & \textbf{23.02} \\
    \bottomrule
\end{tabular}}
\end{table*}

\subsection{Experiment \uppercase\expandafter{\romannumeral2}: Experimental results on CBIS-DDSM and FFDM-DBT dataset}

\begin{figure}[!t]
\centerline{\includegraphics[width=1\linewidth]{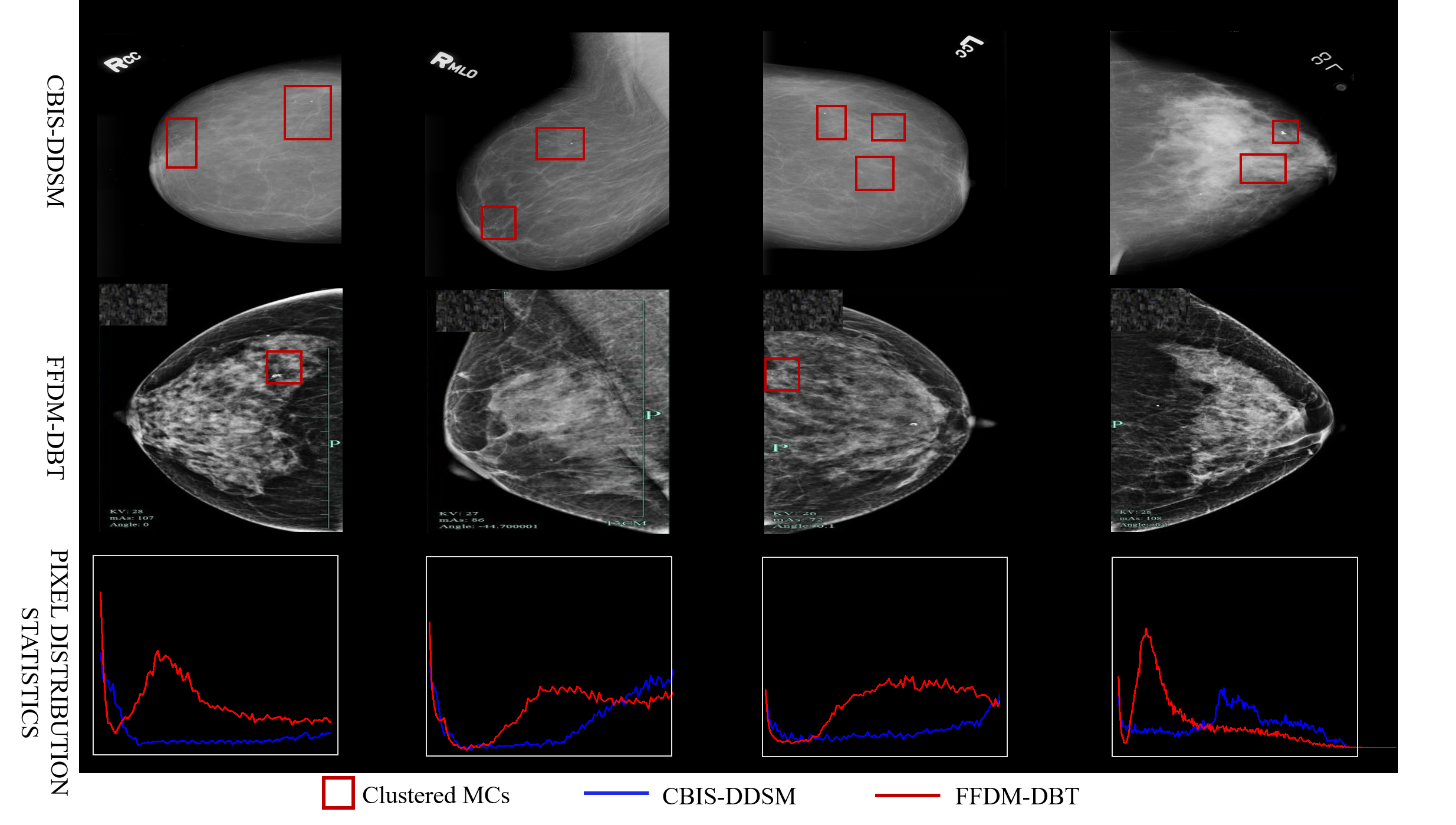}}
\caption{Example cases on FFDM-DBT and CBIS-DDSM datasets. Eight images from CBIS-DDSM and FFDM-DBT datasets is selected to perform a comparative analysis of data variations. This comparison reveals the differences in the pixel mapping approaches used by the two datasets. The third row of this figure presents a quantitative analysis of pixel values for the two types of data. Notably, we excluded from statistical analysis the points with pixel values of 0 and 1, which comprise the background. Additionally, for safeguarding patient privacy, confidential patient information is masked with grey regions on the FFDM-DBT dataset.}
\label{fig7}
\end{figure}

In Experiment \uppercase\expandafter{\romannumeral1}, the efficacy of MLN-net is validated on the FFDM and DBT images which are obtained from the hospital by using different imaging techniques. As shown in Fig. \ref{fig1}, the data retain a significant degree of relevance, despite variations in image acquisition. To further verify the robustness of MLN-net, we select FFDM data from both FFDM-DBT dataset and CBIS-DDSM dataset. The chosen data come from varied hospitals and equipment, resulting in a notable disparity in pixel distribution, as illustrated in Fig. \ref{fig7}. The quantitative results are presented in Table. \ref{tab:2}. And Fig. \ref{fig5} shows the segmentation results of four cases on the target domain CBIS-DDSM. Further details on the results are discussed below.

\subsubsection{Comparison with the basic segmentation methods}
The results from four basic medical image segmentation methods are presented in Table. \ref{tab:2} (P1). These methods' performance has significantly declined. For instance, M-net, which exhibited superior segmentation performance in Experiment \uppercase\expandafter{\romannumeral1}, only achieves TPR of 9.87\% and DSC of 12.32\% in this experimental setting. The method's multi-scale input and output strategy have be confirmed that can improve the segmentation performance under minor domain shifts. However, as the distribution differences in the data domain increase further, the method appears to focus excessively on detailed image features, thereby overlooking the global distribution information. This focus results in an inability to detect clustered microcalcifications. Unet gets DSC of 19.92\% and  HD of 52.51mm on the source domain FFDM-DBT. The DSC and HD decreased by about 25\% and 10mm compared with the results of Table. \ref{tab:1}. Similarly, Swin-Unet demonstrates a significant decrease in segmentation accuracy compared to Experiment \uppercase\expandafter{\romannumeral1}. These results suggest that the robustness and generalization performance of these methods are lacking. Consequently, their generalization quality across all target domains cannot be guaranteed when the domain shift scale is substantial.

\subsubsection{Comparison with the state-of-the-art methods for recognizing clustered microcalcifications}
The Table. \ref{tab:2} (P2) shows the result of state-of-the-art methods for recognizing clustered microcalcifications. Compared with the four basic segmentation methods, the segmentation accuracy of these methods has a significant decline. For instance, CA-net achieves DSC of 33.67\% on the source domain CBIS-DDSM, indicating a decline of approximately 40\% compared to Table. \ref{tab:1}. Moreover, it is noteworthy that Iadml almost loses its ability of clustered microcalcifications segmentation, with DSC of 8.1\%, HD of 77.48mm, and ASD of 52.66mm on the source domain CBIS-DDSM. The poor segmentation effect of Iadml is also revealed in the four examples in Fig. \ref{fig5}. In summary, the overall performance degradation of segmentation involves two key factors. First, CBIS-DDSM dataset demonstrates limitations in effectively characterizing clustered microcalcification lesions, because of the early technological limitations. Second, as shown in Fig. \ref{fig7}, the disparity between CBIS-DDSM and FFDM-DBT datasets is greater than that between FFDM and DBT images within FFDM-DBT dataset. These four current advanced methods for recognizing clustered microcalcifications are lack of robustness to the variations in target distribution. Consequently, it may lead to poor performance, particularly when the dataset exhibits significant distributional shifts.

\subsubsection{Comparison with the state-of-the-art DG methods}
The Table. \ref{tab:2} (P3) presents the results of the the state-of-the-art DG methods. The results reveals that the DG methods improve the generalization performance over the methods of P2 and P1, which largely due to their regularization effect on local learning enabling them to extract general feature representations. For example, BigAug achieves a segmentation accuracy surpassed only by the proposed MLN-net, with DSC of 44.51\%, HD of 33.64mm and ASD of 22.58mm on the source domain CBIS-DDSM, further confirming the positive impact of data augmentation. Similarly, Sadn registers commendable segmentation accuracy. As anticipated, MLN-net outperforms all other methods in terms of average results of DSC, HD, and ASD by significant margins. Specifically, on the source domain CBIS-DDSM, MLN-net achieves the highest average DSC of 50.78\% and ASD of 20.33mm, which are 5.62\% and 2.25mm higher than the suboptimal algorithm, respectively. And as shown in Fig. \ref{fig5}, the proposed MLN-net accurately segments the structure and delineates the boundary in images with unknown distributions, while other DG methods sometimes fall short. These results suggest that MLN-net exhibits superior generalization performance, even under challenging circumstances characterized by significant deviations between the source and target domains.

\section{Discussion}
\subsection{Ablation study}
MLN-net comprises three modules: the source domain data augmentation, the segmentation network, and the branch selection strategy. To demonstrate the effectiveness of each module, ablation studies were presented in FFDM-DBT dataset. And the corresponding experiments with different configurations are given in TABEL \ref{tab:3}, showing some quantitative insight into the performance of each module.

\begin{itemize} \item Basenet: The standard Swin-Unet. \item Basenet-S: The Basenet combined with the source domain data augmentation. \item MLN-net-R: The Basenet combined with the source domain data augmentation and multiple LN layers, and without the branch selection strategy. \item Unet-R: The standard Unet combined with the source domain data augmentation and multiple LN layers, and without the branch selection strategy. \item MLN-net-E: The Basenet combined with the source domain data augmentation, the multiple LN layers and the branch selection strategy based on Euclidean distance. \item Unet-E: The standard Unet combined with the source domain data augmentation, multiple LN layers and the branch selection strategy based on Euclidean distance. \item MLN-net: The Basenet combined with the source domain data augmentation, multiple LN layers and the branch selection strategy based on cosine similarity. \item Unet*: The standard Unet combined with the source domain data augmentation, multiple LN layers and the branch selection strategy based on cosine similarity. \item MLN-net-E+: Compared to MLN-net-E, a correctional mean parameter is added to the matrices $\emph{Q}_{d}$ and $\emph{Q}_{t}$. \item MLN-net+: Compared to MLN-net, a correctional mean parameter is added to the matrices $\emph{Q}_{d}$ and $\emph{Q}_{t}$.
\end{itemize}

\begin{table}[cols=5,pos=h]
  \centering
  \caption{Segmentation performance comparison with different configurations. DSC and HD are used to evaluate the performance of these methods (best result is in bold for each column). Target DBT (Source FFDM): the models are trained on the domain FFDM and tested on the domain DBT, and Target FFDM (Source DBT): the models are trained on the domain DBT and tested on the domain FFDM.}  \label{tab:3}%
  \setlength{\tabcolsep}{1mm}{
    \begin{tabular}{p{5.875em}cccccc}
    \toprule
    Datesets: &       & \multicolumn{2}{c}{Target DBT                                                                                 } &       & \multicolumn{2}{c}{Target FFDM                                                                                                  } \\
     FFDM-DBT &       & \multicolumn{2}{c}{(Source FFDM)} &       & \multicolumn{2}{c}{(Source DBT)} \\
\cmidrule{1-1}\cmidrule{3-4}\cmidrule{6-7}    Method &       & \multicolumn{1}{c}{DSC (\%)} & \multicolumn{1}{c}{HD (mm)} &       & \multicolumn{1}{c}{DSC (\%)} & \multicolumn{1}{c}{HD (mm)} \\
\cmidrule{1-1}\cmidrule{3-4}\cmidrule{6-7}    Basenet &       & 46.97 & 29.17 &       & 57.07 & 26.21 \\
    Basenet-S &       & 67.48 & 25.06 &       & 70.93 & 24.33 \\
    MLN-net-R &       & 68.12 & 25.95 &       & 73.03 & 24.39 \\
    Unet-R &       & 53.62 & 30.10 &       & 53.03 & 31.75 \\
    MLN-net-E &       & 78.76 & 24.02 &       & 84.96 & 22.62 \\
    Unet-E &       & 66.18 & 25.44 &       &66.12 &25.39 \\
    MLN-net &       & 77.91 & 23.32 &       & 85.52 & 20.49 \\
    Unet* &       & 68.99 & 26.47 &       & 69.07 &24.76 \\
    MLN-net-E+ &       & 78.53 & \textbf{22.42} &       & 86.12 & 20.69 \\
    MLN-net+ &       & \textbf{79.12} & 22.75 &       & \textbf{89.37} & \textbf{19.66} \\
\bottomrule
\end{tabular}}

\end{table}%

\subsubsection{Efficacy of the source domain data augmentation method}

\begin{figure}[!t]
\centerline{\includegraphics[width=1\linewidth]{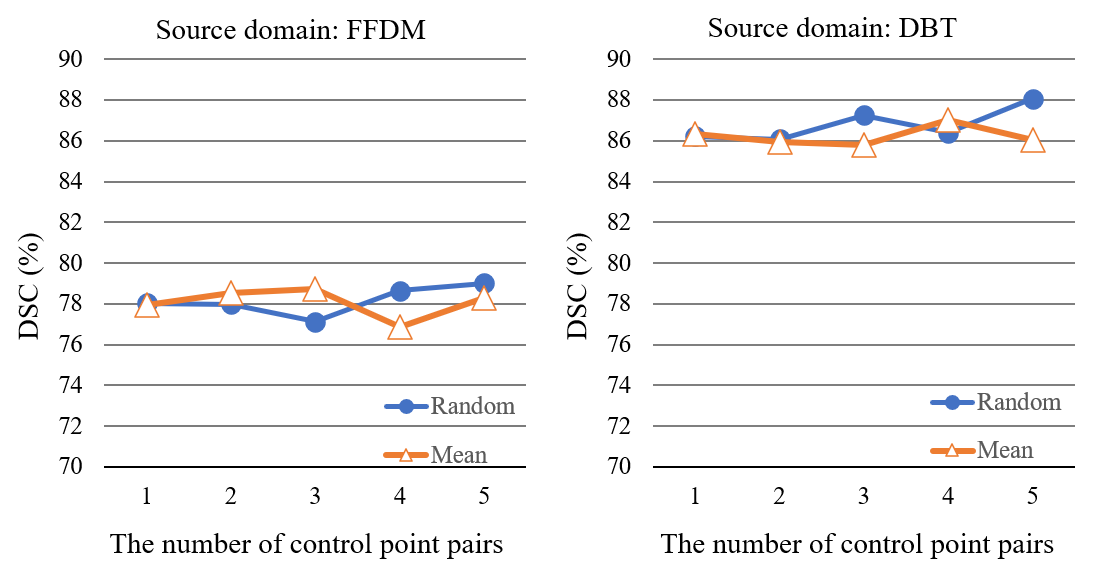}}
\caption{The segmentation performance of MLN-net on FFDM-DBT dataset based on different numbers of control point pairs for the Bézier curve and different selection strategies of parameters. The vertical axis represents the DSC and the horizontal axis represents different number of control point pairs of the Bézier curve.}
\label{fig8}
\end{figure}

The source-similar data augmentation method, based on Bézier curves, requires a manually initialization of hyperparameters. To evaluate the effect of these hyperparameters, an ablation study was carried out, where control point pairs for the Bézier curve were discussed in detail in the context of MLN-net. And the parameters of control point that will be chosen are often set to mean or random values of some variables. The Random strategy takes values arbitrarily chosen between 0 and 0.5 for $a$, while the Mean strategy generates $a$ uniformly, based on the number of control points (i.e., $a_{n}=0.5 \cdot n/N$, where $n$ denotes the label of the control point pair). Once $a$ is established, a control point pair can be identified as ($a$, 1-$a$) and (1-$a$,$a$). Fig. \ref{fig8} represents the results evaluated for DSC, suggesting that the performance of MLN-net is impacted less by either the number of Bézier curve's control point pairs or the selection strategy of parameters, given that they are maintained within a sensible range. However, the increase in the number of control point pairs correspondingly expands the training data volume, resulting in a challenging computational problem. And the Random strategy may introduces uncertainty into MLN-net. As a result, we chose two control points and set $a$ at 0.30 and 0.70 \cite{zhou2022generalizable,zhou2020deep}. And both the start and end points are initialized to (0,0) and (1,1), in accordance with the image's pixel value range.

\begin{figure}[!t]
\centerline{\includegraphics[width=1\linewidth]{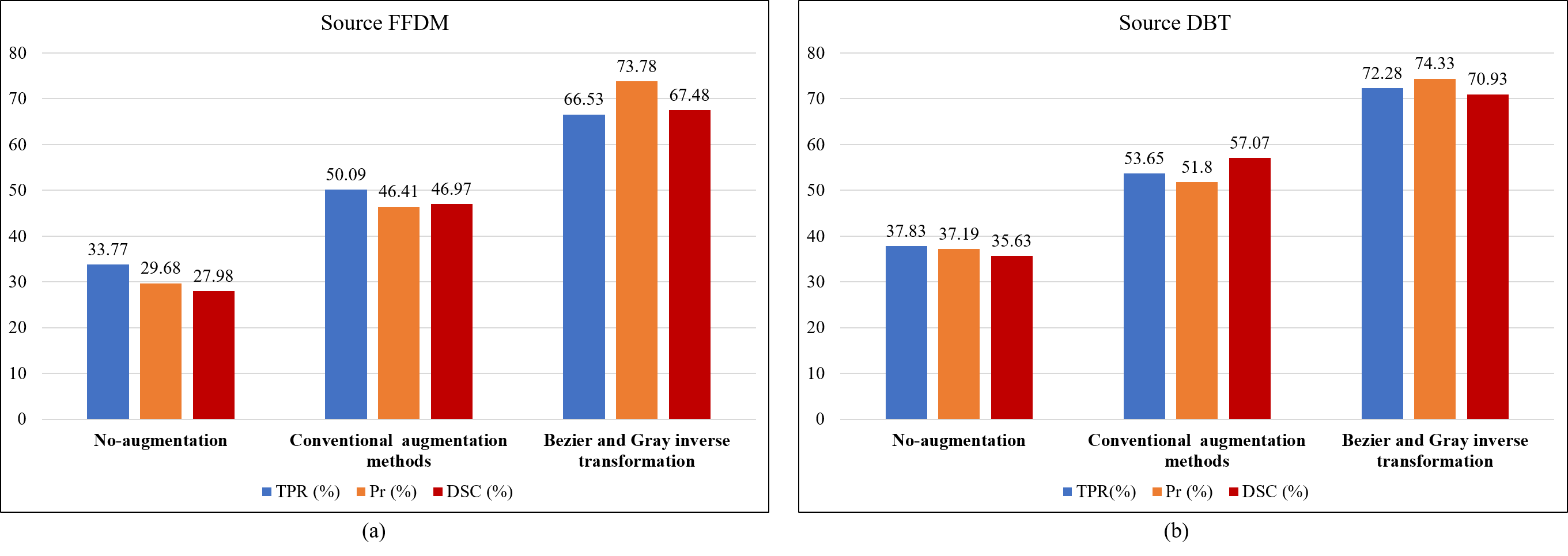}}
\caption{Evaluation comparison of source domain image augmentation method.The experimental results shown in (a) and (b) are from the source domains of FFDM and DBT, respectively.}
\label{fig9}
\end{figure}

Based on these parameters settings, we further investigated the effect of the source domain data augmentation on clustered microcalcifications segmentation. The original Swinunet is selected as the base network for this experiment, to ensure that the results remain untainted by multiple LN layers and varying branch selection strategy. Our control group comprises both non-augmentation and conventional augmentation groups, in which conventional augmentation methods consists of random cropping, flipping, and scaling geometric transformations. Fig. \ref{fig9} presents the experimental results, evaluated by TPR, Pr, and DSC. The results indicate that the use of such source domain data augmentation can lead to significant improvements in generalization. Table. \ref{tab:3} shows that Basenet-S (the addition of the source domain augmentation on Basenet) achieves DSC of 67.48\% and HD of 25.06mm, which presents an significantly improvement compared to the Basenet. It should be noted that Basenet-S does not distinguish between different domain information but amalgamates both source-similar and source-dissimilar data as input. These results also demonstrate the effectiveness of the proposed source domain augmentation module.

\subsubsection{Efficacy of the segmentation network with multiple LN layers and the branch selection strategy}
MLN-net employs the segmentation network with multiple LN layers for capturing image features and domain information and leverages the branch selection strategy to determine optimal segmentation outcomes. This section scrutinizes the benefits of using these modules by comparing the resulting performance as shown in Table \ref{tab:3}. MLN-net-R and Unet-R models do not include the branch selection strategy. The integration over multi-domain outputs can be performed by summing over the pixel values of segmentation results from different branches. The pixel values greater than or equal to 1 are considered as lesion areas.
It can be observed that MLN-net-R (with multiple LN layers structure in Swinunet) can optimize the segmentation results by separating information from different domains. However, it does not yield a significant improvement compared to Basenet-S. MLN-net-E and MLN-net (with branch selection strategies) measure the similarity between target and source data using domain information extracted from normalization layers. 
Combined with branch selection strategies, it allows these models to make optimal decision given all the branch information and ultimately to yield the powerful segmentation algorithm of clustered microcalcifications.

Furthermore, we explored the potential advantages of cosine similarity in the branch selection strategy. As seen in TABEL \ref{tab:3}, when $\emph{Q}_{d}$ and $\emph{Q}_{t}$ only encompass mean and variance of data, the cosine similarity does not exhibit significant superiority over Euclidean distance. The introduction of a correctional mean parameter in $\emph{Q}{d}$ and $\emph{Q}{t}$ provides MLN-net+ an improved segmentation performance on the source domain DBT, with DSC and HD reaching 89.37\% and 19.66mm, respectively. This performance signifies an improvement of 4.41\% and 2.96mm compared with MLN-net-E. Importantly, this correctional mean parameter is taken from the mean of the input pixels, excluding the minimum pixel value that corresponds to the background region in mammography. In practice, the size of the background region fluctuates significantly with respect to various breast sizes, affecting the global pixel mean. The results of MLN-net+ and MLN-net demonstrates the improvements by using additional domain distribution information. Moreover, compared to MLN-Net-E+, MLN-net+ improves DSC and HD from 86.12\% and 20.69mm to 89.37\% and 19.66mm on the source domain DBT, indicating 
the benefits of branch selection strategy with cosine similarity in processing high-dimensional data.

In the ablation studies, we also established baseline models based on Unet framework. We discovered that models based on the Swinunet framework outperform those based on Unet framework with a significant gap. The results verify the superior efficacy of MLN-net's feature extraction network, which is rooted in self-attention mechanisms.

\subsection{Analysis on params of MLN-net}
MLN-net utilizes the Swin-Unet architecture to extract data features and applies multiple LN layers to differentiate the characteristics of various domains, thereby improving the segmentation accuracy of clustered microcalcifications in the presence of target domain shift. However, using multiple LN layers increases the parameters of the segmentation network. To investigate the effects of the multiple LN layers on the network's parameter count, we compared the total number of parameters of Swin-Unet and MLN-net. Both models take single-channel, 512$\times$512 resolution images as input to eliminate the influence of different inputs. The experiment reveals that Swin-Unet and MLN-net have 83.6M and 84.9M parameters, respectively. MLN-net has 1.3M additional parameters compared to Swin-Unet. Nonetheless, Swin-Unet is incapable of addressing domain shift problems, and retraining the network is necessary to achieve accurate segmentation of multi-domain data, which leading to a doubling of the number of required parameters. In contrast, MLN-net effectively performs segmentation tasks on multi-source data with only a small increase in network parameters, providing a new domain shift solution for clustered microcalcifications segmentation.

\section{Conclusion}
In this study, we developed a multi-source medical image segmentation method, MLN-net, for the segmentation of clustered microcalcifications, a key indicator of breast cancer. Combining the source domain data augmentation method based on Bézier curves and grayscale-inversion, it improves data diversity dramatically. And a novel segmentation network is constructed using multiple LN layers, so as to extract features from both source-similar and source-dissimilar data. Furthermore, the branch selection strategy is incorporated into MLN-net by utilizing cosine similarity distance to get the best results for target domain. To our knowledge, MLN-net is the first cross-center generalization method for segmenting clustered microcalcifications in breast cancer diagnosis.

We aimed to develop a clustered microcalcifications segmentation method with cross-center generalization ability, which overcomes the binding characteristics inherent to deep learning models and data sources, thus bridging the data gap between imaging methods of different hospitals. MLN-net, no doubt, provides an innovative solution for intelligent breast cancer detection and treatment, with the potential to be applied widely in a clinically meaningful setting.

While the MLN-net has some important practical properties, it suffers from several limitations particularly in the interpretability. Deep learning-based methods are often difficult to understand the rationale behind the results, posing significant challenges for medical diagnoses based on artificial intelligence and impeding the clinical application of these approaches. Consequently, future research should concentrate on improving the interpretability of MLN-net, introducing the key concepts needed for an understanding of the complex models, and assisting healthcare professionals in disease diagnosis and treatment.

\section{Acknowledgments}
This work is supported by the Zhejiang Provincial Natural Science Foundation of China [No.LQ23F030002], the “Ling Yan” Research and Development Project of Science and Technology Department of Zhejiang Province of China [No.2022C03122], the Public Welfare Technology Application and Research Projects of Science and Technology Department of Zhejiang Province of China [No.LGF21F010004],  and Zhejiang Shuren University Basic Scientific Research Special Funds [2023XZ001].

\printcredits



\begin{thebibliography}{60}
\expandafter\ifx\csname natexlab\endcsname\relax\def\natexlab#1{#1}\fi
\providecommand{\url}[1]{\texttt{#1}}
\providecommand{\href}[2]{#2}
\providecommand{\path}[1]{#1}
\providecommand{\DOIprefix}{doi:}
\providecommand{\ArXivprefix}{arXiv:}
\providecommand{\URLprefix}{URL: }
\providecommand{\Pubmedprefix}{pmid:}
\providecommand{\doi}[1]{\href{http://dx.doi.org/#1}{\path{#1}}}
\providecommand{\Pubmed}[1]{\href{pmid:#1}{\path{#1}}}
\providecommand{\bibinfo}[2]{#2}
\ifx\xfnm\relax \def\xfnm[#1]{\unskip,\space#1}\fi
\bibitem[{Abuduweili et~al.(2021)Abuduweili, Li, Shi, Xu and
  Dou}]{abuduweili2021adaptive}
\bibinfo{author}{Abuduweili, A.}, \bibinfo{author}{Li, X.},
  \bibinfo{author}{Shi, H.}, \bibinfo{author}{Xu, C.Z.}, \bibinfo{author}{Dou,
  D.}, \bibinfo{year}{2021}.
\newblock \bibinfo{title}{Adaptive consistency regularization for
  semi-supervised transfer learning}, in: \bibinfo{booktitle}{Proceedings of
  the IEEE/CVF Conference on Computer Vision and Pattern Recognition}, pp.
  \bibinfo{pages}{6923--6932}.
\bibitem[{Ba et~al.(2016)Ba, Kiros and Hinton}]{ba2016layer}
\bibinfo{author}{Ba, J.L.}, \bibinfo{author}{Kiros, J.R.},
  \bibinfo{author}{Hinton, G.E.}, \bibinfo{year}{2016}.
\newblock \bibinfo{title}{Layer normalization}.
\newblock \bibinfo{journal}{arXiv preprint arXiv:1607.06450} .
\bibitem[{Balaji et~al.(2018)Balaji, Sankaranarayanan and
  Chellappa}]{balaji2018metareg}
\bibinfo{author}{Balaji, Y.}, \bibinfo{author}{Sankaranarayanan, S.},
  \bibinfo{author}{Chellappa, R.}, \bibinfo{year}{2018}.
\newblock \bibinfo{title}{Metareg: Towards domain generalization using
  meta-regularization}.
\newblock \bibinfo{journal}{Advances in neural information processing systems}
  \bibinfo{volume}{31}.
\bibitem[{Bekker et~al.(2015)Bekker, Shalhon, Greenspan and
  Goldberger}]{bekker2015multi}
\bibinfo{author}{Bekker, A.J.}, \bibinfo{author}{Shalhon, M.},
  \bibinfo{author}{Greenspan, H.}, \bibinfo{author}{Goldberger, J.},
  \bibinfo{year}{2015}.
\newblock \bibinfo{title}{Multi-view probabilistic classification of breast
  microcalcifications}.
\newblock \bibinfo{journal}{IEEE Transactions on medical imaging}
  \bibinfo{volume}{35}, \bibinfo{pages}{645--653}.
\bibitem[{Bougourzi et~al.(2023)Bougourzi, Distante, Dornaika and
  Taleb-Ahmed}]{bougourzi2023pdatt}
\bibinfo{author}{Bougourzi, F.}, \bibinfo{author}{Distante, C.},
  \bibinfo{author}{Dornaika, F.}, \bibinfo{author}{Taleb-Ahmed, A.},
  \bibinfo{year}{2023}.
\newblock \bibinfo{title}{Pdatt-unet: Pyramid dual-decoder attention unet for
  covid-19 infection segmentation from ct-scans}.
\newblock \bibinfo{journal}{Medical Image Analysis} , \bibinfo{pages}{102797}.
\bibitem[{Butoi et~al.(2023)Butoi, Ortiz, Ma, Sabuncu, Guttag and
  Dalca}]{butoi2023universeg}
\bibinfo{author}{Butoi, V.I.}, \bibinfo{author}{Ortiz, J.J.G.},
  \bibinfo{author}{Ma, T.}, \bibinfo{author}{Sabuncu, M.R.},
  \bibinfo{author}{Guttag, J.}, \bibinfo{author}{Dalca, A.V.},
  \bibinfo{year}{2023}.
\newblock \bibinfo{title}{Universeg: Universal medical image segmentation}.
\newblock \bibinfo{journal}{arXiv preprint arXiv:2304.06131} .
\bibitem[{Cao et~al.(2023)Cao, Wang, Chen, Jiang, Zhang, Tian and
  Wang}]{cao2023swin}
\bibinfo{author}{Cao, H.}, \bibinfo{author}{Wang, Y.}, \bibinfo{author}{Chen,
  J.}, \bibinfo{author}{Jiang, D.}, \bibinfo{author}{Zhang, X.},
  \bibinfo{author}{Tian, Q.}, \bibinfo{author}{Wang, M.}, \bibinfo{year}{2023}.
\newblock \bibinfo{title}{Swin-unet: Unet-like pure transformer for medical
  image segmentation}, in: \bibinfo{booktitle}{Computer Vision--ECCV 2022
  Workshops: Tel Aviv, Israel, October 23--27, 2022, Proceedings, Part III},
  \bibinfo{organization}{Springer}. pp. \bibinfo{pages}{205--218}.
\bibitem[{Carneiro et~al.(2017)Carneiro, Nascimento and
  Bradley}]{carneiro2017automated}
\bibinfo{author}{Carneiro, G.}, \bibinfo{author}{Nascimento, J.},
  \bibinfo{author}{Bradley, A.P.}, \bibinfo{year}{2017}.
\newblock \bibinfo{title}{Automated analysis of unregistered multi-view
  mammograms with deep learning}.
\newblock \bibinfo{journal}{IEEE transactions on medical imaging}
  \bibinfo{volume}{36}, \bibinfo{pages}{2355--2365}.
\bibitem[{Chen et~al.(2021)Chen, Lu, Yu, Luo, Adeli, Wang, Lu, Yuille and
  Zhou}]{chen2021transunet}
\bibinfo{author}{Chen, J.}, \bibinfo{author}{Lu, Y.}, \bibinfo{author}{Yu, Q.},
  \bibinfo{author}{Luo, X.}, \bibinfo{author}{Adeli, E.},
  \bibinfo{author}{Wang, Y.}, \bibinfo{author}{Lu, L.},
  \bibinfo{author}{Yuille, A.L.}, \bibinfo{author}{Zhou, Y.},
  \bibinfo{year}{2021}.
\newblock \bibinfo{title}{Transunet: Transformers make strong encoders for
  medical image segmentation}.
\newblock \bibinfo{journal}{arXiv preprint arXiv:2102.04306} .
\bibitem[{Chong et~al.(2019)Chong, Weinstein, McDonald and
  Conant}]{chong2019digital}
\bibinfo{author}{Chong, A.}, \bibinfo{author}{Weinstein, S.P.},
  \bibinfo{author}{McDonald, E.S.}, \bibinfo{author}{Conant, E.F.},
  \bibinfo{year}{2019}.
\newblock \bibinfo{title}{Digital breast tomosynthesis: concepts and clinical
  practice}.
\newblock \bibinfo{journal}{Radiology} \bibinfo{volume}{292},
  \bibinfo{pages}{1}.
\bibitem[{Doersch et~al.(2015)Doersch, Gupta and
  Efros}]{doersch2015unsupervised}
\bibinfo{author}{Doersch, C.}, \bibinfo{author}{Gupta, A.},
  \bibinfo{author}{Efros, A.A.}, \bibinfo{year}{2015}.
\newblock \bibinfo{title}{Unsupervised visual representation learning by
  context prediction}, in: \bibinfo{booktitle}{Proceedings of the IEEE
  international conference on computer vision}, pp.
  \bibinfo{pages}{1422--1430}.
\bibitem[{Dong et~al.(2020)Dong, Cong, Sun, Liu and Xu}]{dong2020cscl}
\bibinfo{author}{Dong, J.}, \bibinfo{author}{Cong, Y.}, \bibinfo{author}{Sun,
  G.}, \bibinfo{author}{Liu, Y.}, \bibinfo{author}{Xu, X.},
  \bibinfo{year}{2020}.
\newblock \bibinfo{title}{Cscl: Critical semantic-consistent learning for
  unsupervised domain adaptation}, in: \bibinfo{booktitle}{European Conference
  on Computer Vision}, \bibinfo{organization}{Springer}. pp.
  \bibinfo{pages}{745--762}.
\bibitem[{Dou et~al.(2019)Dou, Coelho~de Castro, Kamnitsas and
  Glocker}]{dou2019domain}
\bibinfo{author}{Dou, Q.}, \bibinfo{author}{Coelho~de Castro, D.},
  \bibinfo{author}{Kamnitsas, K.}, \bibinfo{author}{Glocker, B.},
  \bibinfo{year}{2019}.
\newblock \bibinfo{title}{Domain generalization via model-agnostic learning of
  semantic features}.
\newblock \bibinfo{journal}{Advances in Neural Information Processing Systems}
  \bibinfo{volume}{32}.
\bibitem[{Du et~al.(2019)Du, Tan, Yang, Feng, Xue, Zheng, Ye and
  Zhang}]{du2019ssf}
\bibinfo{author}{Du, L.}, \bibinfo{author}{Tan, J.}, \bibinfo{author}{Yang,
  H.}, \bibinfo{author}{Feng, J.}, \bibinfo{author}{Xue, X.},
  \bibinfo{author}{Zheng, Q.}, \bibinfo{author}{Ye, X.},
  \bibinfo{author}{Zhang, X.}, \bibinfo{year}{2019}.
\newblock \bibinfo{title}{Ssf-dan: Separated semantic feature based domain
  adaptation network for semantic segmentation}, in:
  \bibinfo{booktitle}{Proceedings of the IEEE/CVF International Conference on
  Computer Vision}, pp. \bibinfo{pages}{982--991}.
\bibitem[{Fan et~al.(2021)Fan, Wang, Ke, Yang, Gong and
  Zhou}]{fan2021adversarially}
\bibinfo{author}{Fan, X.}, \bibinfo{author}{Wang, Q.}, \bibinfo{author}{Ke,
  J.}, \bibinfo{author}{Yang, F.}, \bibinfo{author}{Gong, B.},
  \bibinfo{author}{Zhou, M.}, \bibinfo{year}{2021}.
\newblock \bibinfo{title}{Adversarially adaptive normalization for single
  domain generalization}, in: \bibinfo{booktitle}{Proceedings of the IEEE/CVF
  Conference on Computer Vision and Pattern Recognition}, pp.
  \bibinfo{pages}{8208--8217}.
\bibitem[{Fu et~al.(2018)Fu, Cheng, Xu, Wong, Liu and Cao}]{fu2018joint}
\bibinfo{author}{Fu, H.}, \bibinfo{author}{Cheng, J.}, \bibinfo{author}{Xu,
  Y.}, \bibinfo{author}{Wong, D.W.K.}, \bibinfo{author}{Liu, J.},
  \bibinfo{author}{Cao, X.}, \bibinfo{year}{2018}.
\newblock \bibinfo{title}{Joint optic disc and cup segmentation based on
  multi-label deep network and polar transformation}.
\newblock \bibinfo{journal}{IEEE transactions on medical imaging}
  \bibinfo{volume}{37}, \bibinfo{pages}{1597--1605}.
\bibitem[{Giess et~al.(2017)Giess, Pourjabbar, Ip, Lacson, Alper and
  Khorasani}]{giess2017comparing}
\bibinfo{author}{Giess, C.S.}, \bibinfo{author}{Pourjabbar, S.},
  \bibinfo{author}{Ip, I.K.}, \bibinfo{author}{Lacson, R.},
  \bibinfo{author}{Alper, E.}, \bibinfo{author}{Khorasani, R.},
  \bibinfo{year}{2017}.
\newblock \bibinfo{title}{Comparing diagnostic performance of digital breast
  tomosynthesis and full-field digital mammography in a hybrid screening
  environment}.
\newblock \bibinfo{journal}{American Journal of Roentgenology}
  \bibinfo{volume}{209}, \bibinfo{pages}{929--934}.
\bibitem[{Gu et~al.(2020)Gu, Wang, Song, Huang, Aertsen, Deprest, Ourselin,
  Vercauteren and Zhang}]{gu2020net}
\bibinfo{author}{Gu, R.}, \bibinfo{author}{Wang, G.}, \bibinfo{author}{Song,
  T.}, \bibinfo{author}{Huang, R.}, \bibinfo{author}{Aertsen, M.},
  \bibinfo{author}{Deprest, J.}, \bibinfo{author}{Ourselin, S.},
  \bibinfo{author}{Vercauteren, T.}, \bibinfo{author}{Zhang, S.},
  \bibinfo{year}{2020}.
\newblock \bibinfo{title}{Ca-net: Comprehensive attention convolutional neural
  networks for explainable medical image segmentation}.
\newblock \bibinfo{journal}{IEEE transactions on medical imaging}
  \bibinfo{volume}{40}, \bibinfo{pages}{699--711}.
\bibitem[{He et~al.(2016)He, Zhang, Ren and Sun}]{he2016deep}
\bibinfo{author}{He, K.}, \bibinfo{author}{Zhang, X.}, \bibinfo{author}{Ren,
  S.}, \bibinfo{author}{Sun, J.}, \bibinfo{year}{2016}.
\newblock \bibinfo{title}{Deep residual learning for image recognition}, in:
  \bibinfo{booktitle}{Proceedings of the IEEE conference on computer vision and
  pattern recognition}, pp. \bibinfo{pages}{770--778}.
\bibitem[{He et~al.(2022)He, Ying, Zhang and Chu}]{he2022evolutionary}
\bibinfo{author}{He, X.}, \bibinfo{author}{Ying, G.}, \bibinfo{author}{Zhang,
  J.}, \bibinfo{author}{Chu, X.}, \bibinfo{year}{2022}.
\newblock \bibinfo{title}{Evolutionary multi-objective architecture search
  framework: Application to covid-19 3d ct classification}, in:
  \bibinfo{booktitle}{International Conference on Medical Image Computing and
  Computer-Assisted Intervention}, \bibinfo{organization}{Springer}. pp.
  \bibinfo{pages}{560--570}.
\bibitem[{He et~al.(2021)He, Carass, Zuo, Dewey and Prince}]{he2021autoencoder}
\bibinfo{author}{He, Y.}, \bibinfo{author}{Carass, A.}, \bibinfo{author}{Zuo,
  L.}, \bibinfo{author}{Dewey, B.E.}, \bibinfo{author}{Prince, J.L.},
  \bibinfo{year}{2021}.
\newblock \bibinfo{title}{Autoencoder based self-supervised test-time
  adaptation for medical image analysis}.
\newblock \bibinfo{journal}{Medical image analysis} \bibinfo{volume}{72},
  \bibinfo{pages}{102136}.
\bibitem[{Horvat et~al.(2019)Horvat, Keating, Rodrigues-Duarte, Morris and
  Mango}]{horvat2019calcifications}
\bibinfo{author}{Horvat, J.V.}, \bibinfo{author}{Keating, D.M.},
  \bibinfo{author}{Rodrigues-Duarte, H.}, \bibinfo{author}{Morris, E.A.},
  \bibinfo{author}{Mango, V.L.}, \bibinfo{year}{2019}.
\newblock \bibinfo{title}{Calcifications at digital breast tomosynthesis:
  imaging features and biopsy techniques}.
\newblock \bibinfo{journal}{Radiographics} \bibinfo{volume}{39},
  \bibinfo{pages}{307--318}.
\bibitem[{Hossain(2022)}]{hossain2022microc}
\bibinfo{author}{Hossain, M.S.}, \bibinfo{year}{2022}.
\newblock \bibinfo{title}{Microc alcification segmentation using modified u-net
  segmentation network from mammogram images}.
\newblock \bibinfo{journal}{Journal of King Saud University-Computer and
  Information Sciences} \bibinfo{volume}{34}, \bibinfo{pages}{86--94}.
\bibitem[{Ibtehaz and Rahman(2020)}]{ibtehaz2020multiresunet}
\bibinfo{author}{Ibtehaz, N.}, \bibinfo{author}{Rahman, M.S.},
  \bibinfo{year}{2020}.
\newblock \bibinfo{title}{Multiresunet: Rethinking the u-net architecture for
  multimodal biomedical image segmentation}.
\newblock \bibinfo{journal}{Neural networks} \bibinfo{volume}{121},
  \bibinfo{pages}{74--87}.
\bibitem[{In(2019)}]{in2019facts}
\bibinfo{author}{In, T.}, \bibinfo{year}{2019}.
\newblock \bibinfo{title}{Facts \& figures 2019: Us cancer death rate has
  dropped 27\% in 25 years}.
\newblock \bibinfo{journal}{American Cancer} .
\bibitem[{Ioffe and Szegedy(2015)}]{ioffe2015batch}
\bibinfo{author}{Ioffe, S.}, \bibinfo{author}{Szegedy, C.},
  \bibinfo{year}{2015}.
\newblock \bibinfo{title}{Batch normalization: Accelerating deep network
  training by reducing internal covariate shift}, in:
  \bibinfo{booktitle}{International conference on machine learning},
  \bibinfo{organization}{PMLR}. pp. \bibinfo{pages}{448--456}.
\bibitem[{Jakubovitz et~al.(2019)Jakubovitz, Rodrigues and
  Giryes}]{jakubovitz2019lautum}
\bibinfo{author}{Jakubovitz, D.}, \bibinfo{author}{Rodrigues, M.R.},
  \bibinfo{author}{Giryes, R.}, \bibinfo{year}{2019}.
\newblock \bibinfo{title}{Lautum regularization for semi-supervised transfer
  learning}, in: \bibinfo{booktitle}{Proceedings of the IEEE/CVF International
  Conference on Computer Vision Workshops}, pp. \bibinfo{pages}{0--0}.
\bibitem[{Jiang et~al.(2020)Jiang, Huang, Fu, Sun, Lakowski and
  Hu}]{jiang2020deep}
\bibinfo{author}{Jiang, P.}, \bibinfo{author}{Huang, S.}, \bibinfo{author}{Fu,
  Z.}, \bibinfo{author}{Sun, Z.}, \bibinfo{author}{Lakowski, T.M.},
  \bibinfo{author}{Hu, P.}, \bibinfo{year}{2020}.
\newblock \bibinfo{title}{Deep graph embedding for prioritizing synergistic
  anticancer drug combinations}.
\newblock \bibinfo{journal}{Computational and structural biotechnology journal}
  \bibinfo{volume}{18}, \bibinfo{pages}{427--438}.
\bibitem[{Kirillov et~al.(2023)Kirillov, Mintun, Ravi, Mao, Rolland, Gustafson,
  Xiao, Whitehead, Berg, Lo et~al.}]{kirillov2023segment}
\bibinfo{author}{Kirillov, A.}, \bibinfo{author}{Mintun, E.},
  \bibinfo{author}{Ravi, N.}, \bibinfo{author}{Mao, H.},
  \bibinfo{author}{Rolland, C.}, \bibinfo{author}{Gustafson, L.},
  \bibinfo{author}{Xiao, T.}, \bibinfo{author}{Whitehead, S.},
  \bibinfo{author}{Berg, A.C.}, \bibinfo{author}{Lo, W.Y.}, et~al.,
  \bibinfo{year}{2023}.
\newblock \bibinfo{title}{Segment anything}.
\newblock \bibinfo{journal}{arXiv preprint arXiv:2304.02643} .
\bibitem[{Kowald et~al.(2022)Kowald, Barrantes, M{\"o}ller, Palmer,
  Murua~Escobar, Schwerk and Fuellen}]{kowald2022transfer}
\bibinfo{author}{Kowald, A.}, \bibinfo{author}{Barrantes, I.},
  \bibinfo{author}{M{\"o}ller, S.}, \bibinfo{author}{Palmer, D.},
  \bibinfo{author}{Murua~Escobar, H.}, \bibinfo{author}{Schwerk, A.},
  \bibinfo{author}{Fuellen, G.}, \bibinfo{year}{2022}.
\newblock \bibinfo{title}{Transfer learning of clinical outcomes from
  preclinical molecular data, principles and perspectives}.
\newblock \bibinfo{journal}{Briefings in Bioinformatics} \bibinfo{volume}{23},
  \bibinfo{pages}{bbac133}.
\bibitem[{Lee et~al.(2017)Lee, Gimenez, Hoogi, Miyake, Gorovoy and
  Rubin}]{lee2017curated}
\bibinfo{author}{Lee, R.S.}, \bibinfo{author}{Gimenez, F.},
  \bibinfo{author}{Hoogi, A.}, \bibinfo{author}{Miyake, K.K.},
  \bibinfo{author}{Gorovoy, M.}, \bibinfo{author}{Rubin, D.L.},
  \bibinfo{year}{2017}.
\newblock \bibinfo{title}{A curated mammography data set for use in
  computer-aided detection and diagnosis research}.
\newblock \bibinfo{journal}{Scientific data} \bibinfo{volume}{4},
  \bibinfo{pages}{1--9}.
\bibitem[{Li et~al.(2022)Li, Zhang, Zhang, Wang, Ma, Zhang and Wu}]{li2022can}
\bibinfo{author}{Li, Z.}, \bibinfo{author}{Zhang, C.}, \bibinfo{author}{Zhang,
  Y.}, \bibinfo{author}{Wang, X.}, \bibinfo{author}{Ma, X.},
  \bibinfo{author}{Zhang, H.}, \bibinfo{author}{Wu, S.}, \bibinfo{year}{2022}.
\newblock \bibinfo{title}{Can: Context-assisted full attention network for
  brain tissue segmentation}.
\newblock \bibinfo{journal}{Medical Image Analysis} , \bibinfo{pages}{102710}.
\bibitem[{Liu et~al.(2021)Liu, Chen, Qin, Dou and Heng}]{liu2021feddg}
\bibinfo{author}{Liu, Q.}, \bibinfo{author}{Chen, C.}, \bibinfo{author}{Qin,
  J.}, \bibinfo{author}{Dou, Q.}, \bibinfo{author}{Heng, P.A.},
  \bibinfo{year}{2021}.
\newblock \bibinfo{title}{Feddg: Federated domain generalization on medical
  image segmentation via episodic learning in continuous frequency space}, in:
  \bibinfo{booktitle}{Proceedings of the IEEE/CVF Conference on Computer Vision
  and Pattern Recognition}, pp. \bibinfo{pages}{1013--1023}.
\bibitem[{Liu et~al.(2020a)Liu, Dou and Heng}]{liu2020shape}
\bibinfo{author}{Liu, Q.}, \bibinfo{author}{Dou, Q.}, \bibinfo{author}{Heng,
  P.A.}, \bibinfo{year}{2020}a.
\newblock \bibinfo{title}{Shape-aware meta-learning for generalizing prostate
  mri segmentation to unseen domains}, in: \bibinfo{booktitle}{Medical Image
  Computing and Computer Assisted Intervention--MICCAI 2020: 23rd International
  Conference, Lima, Peru, October 4--8, 2020, Proceedings, Part II 23},
  \bibinfo{organization}{Springer}. pp. \bibinfo{pages}{475--485}.
\bibitem[{Liu et~al.(2020b)Liu, Yang, Gao, Liu, Dou, He, Huang, Huang, Luo,
  Zhang et~al.}]{liu2020remove}
\bibinfo{author}{Liu, Z.}, \bibinfo{author}{Yang, X.}, \bibinfo{author}{Gao,
  R.}, \bibinfo{author}{Liu, S.}, \bibinfo{author}{Dou, H.},
  \bibinfo{author}{He, S.}, \bibinfo{author}{Huang, Y.},
  \bibinfo{author}{Huang, Y.}, \bibinfo{author}{Luo, H.},
  \bibinfo{author}{Zhang, Y.}, et~al., \bibinfo{year}{2020}b.
\newblock \bibinfo{title}{Remove appearance shift for ultrasound image
  segmentation via fast and universal style transfer}, in:
  \bibinfo{booktitle}{2020 IEEE 17th International Symposium on Biomedical
  Imaging (ISBI)}, \bibinfo{organization}{IEEE}. pp.
  \bibinfo{pages}{1824--1828}.
\bibitem[{Ma et~al.(2019)Ma, Ji and Gao}]{ma2019neural}
\bibinfo{author}{Ma, C.}, \bibinfo{author}{Ji, Z.}, \bibinfo{author}{Gao, M.},
  \bibinfo{year}{2019}.
\newblock \bibinfo{title}{Neural style transfer improves 3d cardiovascular mr
  image segmentation on inconsistent data}, in: \bibinfo{booktitle}{Medical
  Image Computing and Computer Assisted Intervention--MICCAI 2019: 22nd
  International Conference, Shenzhen, China, October 13--17, 2019, Proceedings,
  Part II 22}, \bibinfo{organization}{Springer}. pp. \bibinfo{pages}{128--136}.
\bibitem[{Mukama et~al.(2020)Mukama, Kharazmi, Xu, Sundquist, Sundquist,
  Brenner and Fallah}]{mukama2020risk}
\bibinfo{author}{Mukama, T.}, \bibinfo{author}{Kharazmi, E.},
  \bibinfo{author}{Xu, X.}, \bibinfo{author}{Sundquist, K.},
  \bibinfo{author}{Sundquist, J.}, \bibinfo{author}{Brenner, H.},
  \bibinfo{author}{Fallah, M.}, \bibinfo{year}{2020}.
\newblock \bibinfo{title}{Risk-adapted starting age of screening for relatives
  of patients with breast cancer}.
\newblock \bibinfo{journal}{JAMA oncology} \bibinfo{volume}{6},
  \bibinfo{pages}{68--74}.
\bibitem[{Pan et~al.(2022)Pan, Song, Zhang, Yang, Zhang, Ji, Zhang, Shi and
  Wang}]{pan2022molecular}
\bibinfo{author}{Pan, W.}, \bibinfo{author}{Song, K.}, \bibinfo{author}{Zhang,
  Y.}, \bibinfo{author}{Yang, C.}, \bibinfo{author}{Zhang, Y.},
  \bibinfo{author}{Ji, F.}, \bibinfo{author}{Zhang, J.}, \bibinfo{author}{Shi,
  J.}, \bibinfo{author}{Wang, K.}, \bibinfo{year}{2022}.
\newblock \bibinfo{title}{The molecular subtypes of triple negative breast
  cancer were defined and a ligand-receptor pair score model was constructed by
  comprehensive analysis of ligand-receptor pairs}.
\newblock \bibinfo{journal}{Frontiers in Immunology} \bibinfo{volume}{13}.
\bibitem[{Perkonigg et~al.(2021)Perkonigg, Hofmanninger, Herold, Brink,
  Pianykh, Prosch and Langs}]{perkonigg2021dynamic}
\bibinfo{author}{Perkonigg, M.}, \bibinfo{author}{Hofmanninger, J.},
  \bibinfo{author}{Herold, C.J.}, \bibinfo{author}{Brink, J.A.},
  \bibinfo{author}{Pianykh, O.}, \bibinfo{author}{Prosch, H.},
  \bibinfo{author}{Langs, G.}, \bibinfo{year}{2021}.
\newblock \bibinfo{title}{Dynamic memory to alleviate catastrophic forgetting
  in continual learning with medical imaging}.
\newblock \bibinfo{journal}{Nature communications} \bibinfo{volume}{12},
  \bibinfo{pages}{5678}.
\bibitem[{Ronneberger et~al.(2015)Ronneberger, Fischer and
  Brox}]{ronneberger2015u}
\bibinfo{author}{Ronneberger, O.}, \bibinfo{author}{Fischer, P.},
  \bibinfo{author}{Brox, T.}, \bibinfo{year}{2015}.
\newblock \bibinfo{title}{U-net: Convolutional networks for biomedical image
  segmentation}, in: \bibinfo{booktitle}{International Conference on Medical
  image computing and computer-assisted intervention},
  \bibinfo{organization}{Springer}. pp. \bibinfo{pages}{234--241}.
\bibitem[{Samala et~al.(2016)Samala, Chan, Hadjiiski, Cha and
  Helvie}]{samala2016deep}
\bibinfo{author}{Samala, R.K.}, \bibinfo{author}{Chan, H.P.},
  \bibinfo{author}{Hadjiiski, L.M.}, \bibinfo{author}{Cha, K.},
  \bibinfo{author}{Helvie, M.A.}, \bibinfo{year}{2016}.
\newblock \bibinfo{title}{Deep-learning convolution neural network for
  computer-aided detection of microcalcifications in digital breast
  tomosynthesis}, in: \bibinfo{booktitle}{Medical Imaging 2016: Computer-Aided
  Diagnosis}, \bibinfo{organization}{SPIE}. pp. \bibinfo{pages}{234--240}.
\bibitem[{Segu et~al.(2023)Segu, Tonioni and Tombari}]{segu2023batch}
\bibinfo{author}{Segu, M.}, \bibinfo{author}{Tonioni, A.},
  \bibinfo{author}{Tombari, F.}, \bibinfo{year}{2023}.
\newblock \bibinfo{title}{Batch normalization embeddings for deep domain
  generalization}.
\newblock \bibinfo{journal}{Pattern Recognition} \bibinfo{volume}{135},
  \bibinfo{pages}{109115}.
\bibitem[{Seo et~al.(2020)Seo, Suh, Kim, Kim, Han and Han}]{seo2020learning}
\bibinfo{author}{Seo, S.}, \bibinfo{author}{Suh, Y.}, \bibinfo{author}{Kim,
  D.}, \bibinfo{author}{Kim, G.}, \bibinfo{author}{Han, J.},
  \bibinfo{author}{Han, B.}, \bibinfo{year}{2020}.
\newblock \bibinfo{title}{Learning to optimize domain specific normalization
  for domain generalization}, in: \bibinfo{booktitle}{European Conference on
  Computer Vision}, \bibinfo{organization}{Springer}. pp.
  \bibinfo{pages}{68--83}.
\bibitem[{Sourati et~al.(2019)Sourati, Gholipour, Dy, Tomas-Fernandez, Kurugol
  and Warfield}]{sourati2019intelligent}
\bibinfo{author}{Sourati, J.}, \bibinfo{author}{Gholipour, A.},
  \bibinfo{author}{Dy, J.G.}, \bibinfo{author}{Tomas-Fernandez, X.},
  \bibinfo{author}{Kurugol, S.}, \bibinfo{author}{Warfield, S.K.},
  \bibinfo{year}{2019}.
\newblock \bibinfo{title}{Intelligent labeling based on fisher information for
  medical image segmentation using deep learning}.
\newblock \bibinfo{journal}{IEEE transactions on medical imaging}
  \bibinfo{volume}{38}, \bibinfo{pages}{2642--2653}.
\bibitem[{Strehl et~al.(2000)Strehl, Ghosh and Mooney}]{strehl2000impact}
\bibinfo{author}{Strehl, A.}, \bibinfo{author}{Ghosh, J.},
  \bibinfo{author}{Mooney, R.}, \bibinfo{year}{2000}.
\newblock \bibinfo{title}{Impact of similarity measures on web-page
  clustering}, in: \bibinfo{booktitle}{Workshop on artificial intelligence for
  web search (AAAI 2000)}, p.~\bibinfo{pages}{64}.
\bibitem[{Sung et~al.(2021)Sung, Ferlay, Siegel, Laversanne, Soerjomataram,
  Jemal and Bray}]{sung2021global}
\bibinfo{author}{Sung, H.}, \bibinfo{author}{Ferlay, J.},
  \bibinfo{author}{Siegel, R.L.}, \bibinfo{author}{Laversanne, M.},
  \bibinfo{author}{Soerjomataram, I.}, \bibinfo{author}{Jemal, A.},
  \bibinfo{author}{Bray, F.}, \bibinfo{year}{2021}.
\newblock \bibinfo{title}{Global cancer statistics 2020: Globocan estimates of
  incidence and mortality worldwide for 36 cancers in 185 countries}.
\newblock \bibinfo{journal}{CA: a cancer journal for clinicians}
  \bibinfo{volume}{71}, \bibinfo{pages}{209--249}.
\bibitem[{Tarver(2012)}]{tarver2012cancer}
\bibinfo{author}{Tarver, T.}, \bibinfo{year}{2012}.
\newblock \bibinfo{title}{Cancer facts \& figures 2012. american cancer society
  (acs) atlanta, ga: American cancer society, 2012. 66 p., pdf. available
  from}.
\bibitem[{Vaswani et~al.(2017)Vaswani, Shazeer, Parmar, Uszkoreit, Jones,
  Gomez, Kaiser and Polosukhin}]{vaswani2017attention}
\bibinfo{author}{Vaswani, A.}, \bibinfo{author}{Shazeer, N.},
  \bibinfo{author}{Parmar, N.}, \bibinfo{author}{Uszkoreit, J.},
  \bibinfo{author}{Jones, L.}, \bibinfo{author}{Gomez, A.N.},
  \bibinfo{author}{Kaiser, {\L}.}, \bibinfo{author}{Polosukhin, I.},
  \bibinfo{year}{2017}.
\newblock \bibinfo{title}{Attention is all you need}.
\newblock \bibinfo{journal}{Advances in neural information processing systems}
  \bibinfo{volume}{30}.
\bibitem[{Wang and Yang(2018)}]{wang2018context}
\bibinfo{author}{Wang, J.}, \bibinfo{author}{Yang, Y.}, \bibinfo{year}{2018}.
\newblock \bibinfo{title}{A context-sensitive deep learning approach for
  microcalcification detection in mammograms}.
\newblock \bibinfo{journal}{Pattern recognition} \bibinfo{volume}{78},
  \bibinfo{pages}{12--22}.
\bibitem[{Wang et~al.(2020)Wang, Yu, Li, Yang, Fu and Heng}]{wang2020dofe}
\bibinfo{author}{Wang, S.}, \bibinfo{author}{Yu, L.}, \bibinfo{author}{Li, K.},
  \bibinfo{author}{Yang, X.}, \bibinfo{author}{Fu, C.W.},
  \bibinfo{author}{Heng, P.A.}, \bibinfo{year}{2020}.
\newblock \bibinfo{title}{Dofe: Domain-oriented feature embedding for
  generalizable fundus image segmentation on unseen datasets}.
\newblock \bibinfo{journal}{IEEE Transactions on Medical Imaging}
  \bibinfo{volume}{39}, \bibinfo{pages}{4237--4248}.
\bibitem[{Wang et~al.(2021)Wang, Nayak, Guttery, Zhang and
  Zhang}]{wang2021covid}
\bibinfo{author}{Wang, S.H.}, \bibinfo{author}{Nayak, D.R.},
  \bibinfo{author}{Guttery, D.S.}, \bibinfo{author}{Zhang, X.},
  \bibinfo{author}{Zhang, Y.D.}, \bibinfo{year}{2021}.
\newblock \bibinfo{title}{Covid-19 classification by ccshnet with deep fusion
  using transfer learning and discriminant correlation analysis}.
\newblock \bibinfo{journal}{Information Fusion} \bibinfo{volume}{68},
  \bibinfo{pages}{131--148}.
\bibitem[{Wei et~al.(2019)Wei, Meng, Zhao, Xu and Wu}]{wei2019semi}
\bibinfo{author}{Wei, W.}, \bibinfo{author}{Meng, D.}, \bibinfo{author}{Zhao,
  Q.}, \bibinfo{author}{Xu, Z.}, \bibinfo{author}{Wu, Y.},
  \bibinfo{year}{2019}.
\newblock \bibinfo{title}{Semi-supervised transfer learning for image rain
  removal}, in: \bibinfo{booktitle}{Proceedings of the IEEE/CVF conference on
  computer vision and pattern recognition}, pp. \bibinfo{pages}{3877--3886}.
\bibitem[{Wichakam et~al.(2018)Wichakam, Chayakulkheeree and
  Vateekul}]{wichakam2018deep}
\bibinfo{author}{Wichakam, I.}, \bibinfo{author}{Chayakulkheeree, J.},
  \bibinfo{author}{Vateekul, P.}, \bibinfo{year}{2018}.
\newblock \bibinfo{title}{Deep multi-label 3d convnet for breast cancer
  diagnosis in dbt with inversion augmentation}, in: \bibinfo{booktitle}{Tenth
  international conference on digital image processing (ICDIP 2018)},
  \bibinfo{organization}{SPIE}. pp. \bibinfo{pages}{1565--1577}.
\bibitem[{Youlden et~al.(2012)Youlden, Cramb, Dunn, Muller, Pyke and
  Baade}]{youlden2012descriptive}
\bibinfo{author}{Youlden, D.R.}, \bibinfo{author}{Cramb, S.M.},
  \bibinfo{author}{Dunn, N.A.}, \bibinfo{author}{Muller, J.M.},
  \bibinfo{author}{Pyke, C.M.}, \bibinfo{author}{Baade, P.D.},
  \bibinfo{year}{2012}.
\newblock \bibinfo{title}{The descriptive epidemiology of female breast cancer:
  an international comparison of screening, incidence, survival and mortality}.
\newblock \bibinfo{journal}{Cancer epidemiology} \bibinfo{volume}{36},
  \bibinfo{pages}{237--248}.
\bibitem[{Zhang et~al.(2021)Zhang, Wang, Hou, Wu, Wang, Okumura and
  Shinozaki}]{zhang2021flexmatch}
\bibinfo{author}{Zhang, B.}, \bibinfo{author}{Wang, Y.}, \bibinfo{author}{Hou,
  W.}, \bibinfo{author}{Wu, H.}, \bibinfo{author}{Wang, J.},
  \bibinfo{author}{Okumura, M.}, \bibinfo{author}{Shinozaki, T.},
  \bibinfo{year}{2021}.
\newblock \bibinfo{title}{Flexmatch: Boosting semi-supervised learning with
  curriculum pseudo labeling}.
\newblock \bibinfo{journal}{Advances in Neural Information Processing Systems}
  \bibinfo{volume}{34}, \bibinfo{pages}{18408--18419}.
\bibitem[{Zhang et~al.(2022)Zhang, Karanth, Patel, Murphy and
  Jiang}]{zhang2022multi}
\bibinfo{author}{Zhang, K.}, \bibinfo{author}{Karanth, S.},
  \bibinfo{author}{Patel, B.}, \bibinfo{author}{Murphy, R.},
  \bibinfo{author}{Jiang, X.}, \bibinfo{year}{2022}.
\newblock \bibinfo{title}{A multi-task gaussian process self-attention neural
  network for real-time prediction of the need for mechanical ventilators in
  covid-19 patients}.
\newblock \bibinfo{journal}{Journal of Biomedical Informatics}
  \bibinfo{volume}{130}, \bibinfo{pages}{104079}.
\bibitem[{Zhang et~al.(2020)Zhang, Wang, Yang, Sanford, Harmon, Turkbey, Wood,
  Roth, Myronenko, Xu et~al.}]{zhang2020generalizing}
\bibinfo{author}{Zhang, L.}, \bibinfo{author}{Wang, X.}, \bibinfo{author}{Yang,
  D.}, \bibinfo{author}{Sanford, T.}, \bibinfo{author}{Harmon, S.},
  \bibinfo{author}{Turkbey, B.}, \bibinfo{author}{Wood, B.J.},
  \bibinfo{author}{Roth, H.}, \bibinfo{author}{Myronenko, A.},
  \bibinfo{author}{Xu, D.}, et~al., \bibinfo{year}{2020}.
\newblock \bibinfo{title}{Generalizing deep learning for medical image
  segmentation to unseen domains via deep stacked transformation}.
\newblock \bibinfo{journal}{IEEE transactions on medical imaging}
  \bibinfo{volume}{39}, \bibinfo{pages}{2531--2540}.
\bibitem[{Zheng et~al.(2020)Zheng, Sun, Wu, Jiang, Peng, Yang, Zhang and
  Li}]{zheng20203d}
\bibinfo{author}{Zheng, J.}, \bibinfo{author}{Sun, H.}, \bibinfo{author}{Wu,
  S.}, \bibinfo{author}{Jiang, K.}, \bibinfo{author}{Peng, Y.},
  \bibinfo{author}{Yang, X.}, \bibinfo{author}{Zhang, F.}, \bibinfo{author}{Li,
  M.}, \bibinfo{year}{2020}.
\newblock \bibinfo{title}{3d context-aware convolutional neural network for
  false positive reduction in clustered microcalcifications detection}.
\newblock \bibinfo{journal}{IEEE Journal of Biomedical and Health Informatics}
  \bibinfo{volume}{25}, \bibinfo{pages}{764--773}.
\bibitem[{Zhou et~al.(2020)Zhou, Yang, Hospedales and Xiang}]{zhou2020deep}
\bibinfo{author}{Zhou, K.}, \bibinfo{author}{Yang, Y.},
  \bibinfo{author}{Hospedales, T.}, \bibinfo{author}{Xiang, T.},
  \bibinfo{year}{2020}.
\newblock \bibinfo{title}{Deep domain-adversarial image generation for domain
  generalisation}, in: \bibinfo{booktitle}{Proceedings of the AAAI Conference
  on Artificial Intelligence}, pp. \bibinfo{pages}{13025--13032}.
\bibitem[{Zhou et~al.(2022)Zhou, Qi, Yang, Ni and Shi}]{zhou2022generalizable}
\bibinfo{author}{Zhou, Z.}, \bibinfo{author}{Qi, L.}, \bibinfo{author}{Yang,
  X.}, \bibinfo{author}{Ni, D.}, \bibinfo{author}{Shi, Y.},
  \bibinfo{year}{2022}.
\newblock \bibinfo{title}{Generalizable cross-modality medical image
  segmentation via style augmentation and dual normalization}, in:
  \bibinfo{booktitle}{Proceedings of the IEEE/CVF Conference on Computer Vision
  and Pattern Recognition}, pp. \bibinfo{pages}{20856--20865}.

\end{thebibliography}

\end{document}